\documentclass[a4paper,fleqn]{cas-sc} 

\usepackage[numbers]{natbib}
\usepackage{algorithm}
\usepackage{algpseudocode}
\usepackage{subcaption}
\usepackage{graphicx}
\usepackage{placeins}

\begin{document}
\let\WriteBookmarks\relax
\def\floatpagepagefraction{1}
\def\textpagefraction{.001}

\shorttitle{}    

\shortauthors{}  

\title [mode = title]{Spatio-Temporal Prediction of Unsteady Airfoil Aerodynamics Using Augmented Graph Neural Ordinary Differential Equations with Exogenous Controls}

\author[1]{Henrik Lange}[orcid=0009-0002-9295-7346,
                         linkedin=henrik-lange-867625211]
\cormark[1]
\ead{h.lange@dlr.de}
\credit{Conceptualization, Methodology, Software, Validation, Formal analysis, Investigation, Data Curation, Writing - Original, Writing - Review \& Editing, Visualization}

\author[2]{Reik Thormann}
\ead{reik.thormann@airbus.com}
\credit{Conceptualization, Writing - Review \& Editing, Supervision, Project administration}

\author[1]{Philipp Bekemeyer}[orcid=0009-0001-9888-2499]
\ead{philipp.bekemeyer@dlr.de}
\credit{Conceptualization, Resources, Writing - Review \& Editing, Supervision, Project administration, Funding acquisition}

\affiliation[1]{organization={German Aerospace Center (DLR)},
                addressline={Lilienthalpl. 7}, 
                postcode={38108}, 
                city={Brunswick},
                postcodesep={},
                country={Germany}}

\affiliation[2]{organization={Airbus Operations},
                addressline={Airbus Allee 1}, 
                postcode={28199}, 
                city={Bremen},
                postcodesep={},
                country={Germany}}

\cortext[1]{Corresponding author}

\begin{abstract}
Unsteady aerodynamic phenomena, such as gusts, turbulence, and fluid-structure interactions affect an aircraft during flight. For design, optimisation and certification, it is indispensable to quantify such unsteady aerodynamic effects.
Industry-standard computational fluid dynamics methods, such as solving the unsteady Reynolds-averaged Navier-Stokes equations or the linearized frequency domain method, are either computationally expensive or restricted by assumptions like linearity. Once trained, machine learning methods are capable of computing non-linear relationships very fast, making them suitable as surrogate models. By autoregressively applying graph neural networks (GNNs), operating on a discretised spatial domain, spatio-temporal predictions can be made.
However, autoregressive GNNs suffer from error accumulation leading to unstable rollouts over time.
Here we show that combining GNNs with augmented Neural Ordinary Differential Equations yields temporally stable predictions of the surface forces on a pitching airfoil.
We found that our approach, called GNODE, based on Graph Neural Ordinary Differential Equations, provides temporally more stable, spatially smoother, and overall more accurate results than an autoregressive GNN baseline. Tests are conducted on a dataset consisting of a simulations of a pitching airfoil, including transonic shocks, transient behaviour and dynamic non-linearities. Augmenting GNODEs with additional latent dimensions improves the expressivity and accuracy by capturing underlying history effects.
The developed method demonstrates an approach that is suitable to model non-linear spatio-temporal systems with exogenous inputs.
\nocite{*}
\end{abstract}


\begin{highlights}
\item The graph-based, continuous-time GNODE framework predicts unsteady transonic airfoil aerodynamics.
\item Augmentation helps capturing history effects and reduces aerodynamic phase lag.
\item The model outperforms a time-discrete autoregressive baseline in accuracy and stability.
\end{highlights}


\maketitle

\section{Introduction}\label{sec:introduction}
Design, optimisation, and analysis of aircraft relies on \textit{Computational Fluid Dynamics} (CFD) to compute flow solutions at various flight conditions. Accurate predictions of transonic flow phenomena including discontinuities such as shocks, are mandatory for the application to commercial aircraft. While many scenarios can be sufficiently analysed using steady-state simulations, time-accurate flow solutions must be considered for aeroelastic, stability, and gust load analysis \cite{easa2023certification}. However, resolving time-accurate aerodynamic phenomena, using industry-standard high-fidelity solvers that solve the \textit{Unsteady Reynolds-Averaged Navier-Stokes} (URANS) equations, introduces severe computational burdens. This problem is even more pronounced if such a time-accurate flow solver is iteratively coupled with a structural solver to analyse aeroelastic behaviour. Whereas such simulations are possible at single flow conditions, scenarios that require a large number of time-accurate CFD simulations, e.g. optimisation, are unfeasible due to their computational cost. Consequently, it is of interest to develop faster, yet accurate, approaches.

To achieve this speed-up, the field of \textit{Reduced Order Modelling} (ROM) offers various approaches and has been an active field of aerodynamics and aeroelastics research. Solving the linearised flow equations in the frequency domain (LFD) reduces computational cost by more than an order of magnitude compared to URANS \cite{thormann2013linear}. However, this approach is only valid for small perturbations. In addition to LFD, which simplifies the governing equations, a large field of data-driven approaches has previously been investigated. Among these approaches \textit{Proper Orthogonal Decomposition} with interpolation (POD+I) is widely used to construct a low-dimensional subspace based on the dominant spatial modes of a flow field \cite{weiss2019tutorial, bui2003proper} which can afterwards be exploited for efficient latent-space interpolation. Similarly, \textit{Dynamic Mode Decomposition} (DMD) approximates a flow field using distinct frequencies and growth rates of spatio-temporal coherent features that are extracted from flow snapshots \cite{schmid2010dynamic, kutz2016dynamic}. In addition, system identification methods such as the \textit{Eigensystem Realisation Algorithm} (ERA) construct discrete-time linear state-space models from input-output data \cite{brunton2014state}. To account for weakly non-linear effects, \textit{Volterra series expansions} approximate the aerodynamic response using non-linear kernels \cite{silva1993application}. Another low-dimensional approach is offered by the \textit{Sparse Identification of Non-linear Dynamics} (SINDy) algorithm, which describes a spatio-temporal system using a sparse set of non-linear functions chosen from a candidate library \cite{brunton2016discovering}. However, scaling SINDy to high-dimensional, unstructured fluid domains remains a challenge.

Despite extensive modelling efforts these traditional ROMs are restricted to dynamically linear or weakly non-linear behaviour. However, many future aircraft designs are envisioned to incorporate concepts like flexible high-aspect-ratio wings that are expected to be severely affected by dynamically non-linear behaviour like limit cycle oscillations (LCOs) with strong shock movements \cite{stickan2018explanation}. In addition, such configurations are seen to be more strongly affected by large amplitude gusts which also encompass dynamic non-linearities \cite{friedewald2025gust}. Due to the limitations of traditional ROMs, computationally expensive URANS simulations are state-of-the-art to accurately capture these dynamically non-linear effects. To fill this technology gap, the application of \textit{Machine Learning} (ML) methods to fluid dynamics is an active field of research \cite{brunton2020machine}. \textit{Neural Networks} (NNs) are the main building block of these ML models. As universal function approximators \cite{hornik1989multilayer}, they provide a theoretical foundation for highly non-linear problems. Such models may be used to predict scalar values, e.g. integral aerodynamic quantities like lift, drag, or pitching moment. Even though this is sufficient for many applications, in this work we are interested in distributed quantities, i.e. flow fields. Most basically a \textit{Fully-Connected Neural Network} (FCNN) can be applied point-wise to predict fields \cite{sabater2022fast}. However, FCNNs do not take into account any spatial relationships. To overcome this, \textit{Convolutional Neural Networks} (CNNs) apply spatial limitations on in a Cartesian-structured spatial domain \cite{bhatnagar2019prediction}. However, in aerodynamic applications with an unstructured discretisation, CNNs are not directly applicable. \textit{Graph Neural Networks} (GNNs) overcome this limitation by defining the spatial domain as a graph and performing computations along this structure \cite{battaglia2018relational}. Recent advancements show good results for different types of GNNs for the prediction of steady-state aerodynamics \cite{bronstein2017geometric, hines2023graph, hines2026prediction, harsch2021direct}. Additionally, Neural Operators are a competing state-of-the-art approach that approximates the flow field in a continuous spatial domain instead of using discrete points \cite{serrano2023operator, serrano2023infinity, catalani2024neural}.

To advance ML approaches from steady to unsteady aerodynamic problems, special attention must be given to the temporal dimension. Therefore, approaches tailored to time-series data such as \textit{Recurrent Neural Networks} (RNNs) and specific variations like \textit{Gated Recurrent Units} (GRUs) or \textit{Long Short-Term Memory} (LSTMs) are available \cite{cho2014learning, hochreiter1997long}. While such models are used to predict scalar values, the high dimensionality of flow fields often render them unsuitable \cite{zahn2021application, pan2024improved, ribeiro2023unsteady}. More recently attention-based approaches \cite{vaswani2017attention} replaced traditional RNNs as they can be processed in parallel and capture very long-term dependencies. However, the quadratic complexity of the mechanism can lead to high memory usage. Various works apply the above-mentioned temporal models in a latent space using dimensionality reduction methods like POD or \textit{Autoencoders} (AEs) \cite{lecun2015deep}. These hybrid models enable spatio-temporal predictions at the expense of multiple model components that must be considered \cite{han2022predicting, massegur2024recurrent}. In addition to hybrid approaches GNNs might be used for spatio-temporal problems by applying them in an autoregressive manner \cite{sanchez2020learning, pfaff2020learning}. While such approaches have a strong capability of capturing spatial dependencies through the graph structure, the temporal evolution is simplistic and suffers from autoregressive error accumulation. To overcome this \textit{Neural Ordinary Differential Equations} (Neural ODEs) apply NNs with more sophisticated temporal schemes \cite{chen2018neural}. Adding augmented dimensions improves the expressivity of Neural ODEs \cite{dupont2019augmented}. The combination of GNNs with Neural ODEs results in a spatio-temporal model with a continuous temporal and discretised spatial dimension \cite{poli2019graph}. In \cite{colombo2025data} this approach is used to predict aeroelastic forces. Yet, its extension to full unsteady flow fields under dynamic exogenous inputs remains unexplored.

This work develops a model that combines a GNN to capture spatial dependencies with the Neural ODE approach for the temporal dimension and can handle exogenous inputs. The resulting model is tested on a dataset consisting of unsteady simulations of a two-dimensional airfoil that undergoes a forced pitching motion. A comparison to an autoregressive GNN is also discussed. To the best of the authors' knowledge this work adds the following contributions to the current state of research:
\begin{itemize}
    \item \textbf{Formulation of a continuous-time geometric deep learning framework for unsteady aerodynamics:} Graph Neural Ordinary Differential Equations (GNODEs), combining Graph Neural Networks and Neural ODEs, are introduced to model the continuous-time evolution of spatio-temporal aerodynamic fields subjected to dynamic, exogenous control inputs.
    \item \textbf{Benchmark against discrete-time baseline:} the GNODE framework is systematically evaluated against a state-of-the-art discrete-time autoregressive Graph Network Simulator (GNS) baseline, demonstrating superior stability and shock-capturing capabilities under highly non-linear transonic flight conditions.
    \item \textbf{Analysis of state augmentation:} an ablation study is performed to isolate the role of latent dimensions, demonstrating that state-space augmentation enhances the expressivity of the framework and reduces the phase-lag between predictions and reference data.
\end{itemize}

The remainder of this paper is structured as follows: 
Section \ref{sec:problemDescription} formalises the problem description of this work by first outlining the task of spatio-temporal modelling with exogenous inputs on a high, generic level, followed by the specific task description of modelling the unsteady aerodynamics of an airfoil under forced motion. Subsequently, in Section \ref{sec:methods} the proposed frameworks are discussed: an autoregressive GNNs is employed as a baseline approach. Based on Neural ODEs, Graph Neural ODEs are explained. All methods are compared using a high-fidelity dataset that is introduced in Section \ref{sec:data} and consists of URANS simulations of an airfoil undergoing sinusoidal motions. In Section \ref{sec:results} the discrete-time autoregressive GNS and the continuous-time GNODE frameworks are evaluated and compared using several metrics and visualisations. An ablation study regarding augmentation is performed. Finally, Section \ref{sec:conclusions} outlines the findings of this work, states limitations of the proposed frameworks, and presents future research directions.

\section{Problem Description}\label{sec:problemDescription}
This paper investigates the unsteady aerodynamics of an airfoil undergoing prescribed motion. To establish a rigorous foundation, the problem is first formalised as a spatio-temporal dynamical system with exogenous inputs. Spatio-temporal dynamics describe systems that evolve simultaneously across a temporal domain $\mathcal{T} \subseteq \mathbb{R}$ and a spatial domain $\Omega \subseteq \mathbb{R}^{d_{\text{spatial}}}$. The state of the system at time $t \in [0,T]$ and spatial location $\mathbf{x} \in \Omega$ is denoted as $\mathbf{y}(\mathbf{x}, t)$. At any fixed time $t$, the full spatial state is denoted as $\mathbf{Y}_t = \mathbf{Y}(\cdot, t)$, which resides in a function space $\mathcal{V}$ defined over $\Omega$. In an autonomous system, evolution is governed solely by the internal physics. Consequently, the state at time $t$ is uniquely determined by the initial condition $\mathbf{Y}_0$:
\begin{equation}\label{eq:autonomousSystem}
    \mathbf{Y}_t = \mathcal{S}_{\text{autonomous}}(\mathbf{Y}_0, t),
\end{equation}
where $\mathcal{S}_{\text{autonomous}}$ is the autonomous evolution operator. However, this work focuses on non-autonomous systems subject to time-varying exogenous inputs (forcing terms) $\mathbf{u}(\mathbf{x}, t)$. The spatially distributed input at time $t$, denoted $\mathbf{U}_t = \mathbf{U}(\cdot, t)$, belongs to an input function space $\mathcal{U}$. In such systems, the state $\mathbf{Y}_t$ depends on both the initial condition and the history of exogenous inputs up to time $t$, denoted $\mathbf{U}_{[0,t]} = \{\mathbf{U}_\tau \mid 0 \leq \tau \leq t\}$. The evolution is expressed as:
\begin{equation}\label{eq:nonAutonomousSystem}
    \mathbf{Y}_t = \mathcal{S}_{\text{non-autonomous}}(\mathbf{Y}_0, \mathbf{U}_{[0,t]}, t),
\end{equation}
where $\mathcal{S}_{\text{non-autonomous}}$ is the non-autonomous evolution operator. The abstract objective of this work is to identify a learnable surrogate model $\mathcal{M}_\theta$, parametrised by $\theta$, that approximates $\mathcal{S}_{\text{non-autonomous}}$ and maps the initial state and exogenous input trajectory to the resulting state $\mathbf{Y}_t$.

\begin{figure}
    \centering
    \includegraphics[width=0.55\textwidth]{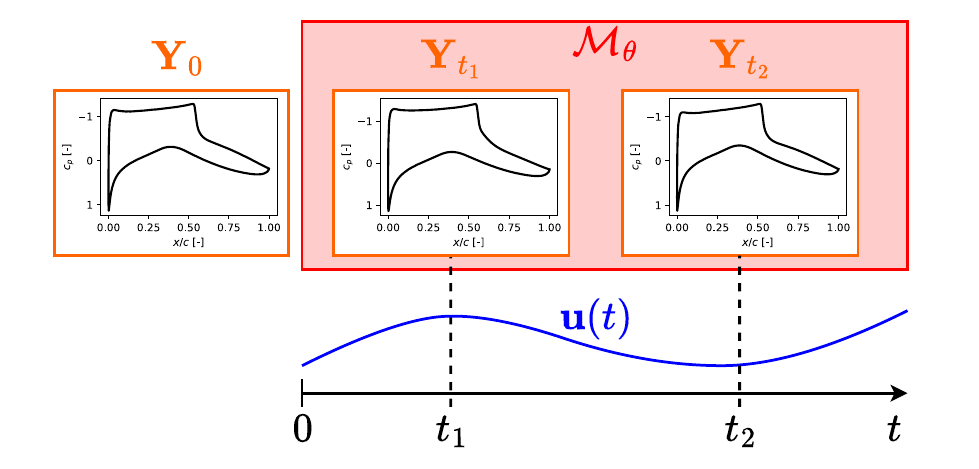}
    \caption{Visualisation of the modelling task (see Equation \ref{eq:learningTask}).}
    \label{fig:learningTask}
\end{figure}

The unsteady flow field around an airfoil undergoing prescribed motion constitutes a special case of the abstract
description of a non-autonomous system. In this context, exogenous controls that act on the system describe the rigid-body movement
of the airfoil. Consequently, the distributed field $\mathbf{u}(\mathbf{x},t)$ reduces to a lower-dimensional control vector $\mathbf{u}_t \in \mathbb{R}^c$, where $c$ denotes the number of independent motion parameters, e.g. pitch angle and angular velocity. For CFD, the spatial domain $\Omega$ is discretised using a mesh of $n$ nodes. The flow field at time $t$ is represented as a snapshot matrix $\mathbf{Y}_t \in \mathbb{R}^{n,q}$, where $q$ is the number of flow variables per node. If the area of interest is restricted to aerodynamic forces on the airfoil surface, the problem can be further reduced to the $n_{\text{surface}}$ boundary nodes, yielding $\mathbf{Y}_{t,\text{surface}} \in \mathbb{R}^{n_{\text{surface}},q}$. Given an initial flow solution $\mathbf{Y}_0$ and the sequence of control inputs $\mathbf{u}_{[0,t]}$, the goal is to learn surrogate model $\mathcal{M}_\theta$ such that:
\begin{equation}\label{eq:learningTask}
    \mathbf{Y}_t = \mathcal{M}_\theta(\mathbf{Y}_0, \mathbf{u}_{[0,t]}, t).
\end{equation}
Figure \ref{fig:learningTask} illustrates the supervised learning objective defined in Equation \ref{eq:learningTask}.

\section{Methods}\label{sec:methods}
This Section outlines the algorithmic framework of the proposed methodology and the baseline method that is used for comparison. First Graph Neural Networks and Neural Ordinary Differential Equations are introduced followed by the Graph Neural Ordinary Differential Equations framework proposed herin.

\subsection{Graph Neural Networks}\label{subsec:graphNetworkSimulator}
Graph Neural Networks are the core building block of the newly proposed GNODE framework as well as for the baseline autoregressive formulation. Spatially discretised data can be represented as a graph $\mathcal{G}=(V,\mathbf{A})$, where $V$ is a set of $n$ nodes $v_i$. The $n$ graph nodes correspond to CFD mesh nodes in Section \ref{sec:problemDescription}. The adjacency matrix $\mathbf{A}\in\mathbb{R}^{n,n}$ defines through $m$ binary entries whether two neighbouring nodes $v_i$ and $v_j$ are connected by a directed edge $e_{j,i}$. The set of neighbours of node $v_i$ is defined as $v_j\in\mathcal{N}(v_i)$. By adding additional information as $d_\text{node}$ node features $\mathbf{x}_i\in\mathbb{R}^{d_\text{node}}$ to each node, and $d_\text{edge}$ edge features $\mathbf{e}_{j,i}\in\mathbb{R}^{d_\text{edge}}$ to each edge the purely structural graph becomes an attributed graph $\mathcal{G}=(V,\mathbf{A},\mathbf{X},\mathbf{E})$, where node and edge features are collected in respective matrices $\mathbf{X}\in\mathbb{R}^{n,d_\text{node}}$ and $\mathbf{E}\in\mathbb{R}^{m,d_\text{edge}}$.

Graph Neural Networks (GNN) are ML architectures that operate on such graph-structured data. The structural information of the graph is a relational inductive bias which can enhance a model \cite{battaglia2018relational}. The Graph Network Simulator (GNS) is introduced in \cite{sanchez2020learning} and refined in \cite{pfaff2020learning} as an autoregressive surrogate for mesh-based simulations, e.g. CFD. For this study the method is adapted to handle control inputs and used as a baseline comparison. GNS is constructed in an encoder-processor-decoder structure. First, a multilayer perceptron (MLP) $\text{MLP}_\text{encode}$ produces an $d_l$-dimensional encoded representation $\mathbf{h}_i^0\in\mathbb{R}^{d_l}$ of the node features $\mathbf{x}_i^\text{input}$ of an attributed graph $\mathcal{G}=(V,\mathbf{A},\mathbf{X},\mathbf{E})$:
\begin{equation}\label{eq:encoder}
    \mathbf{h}_i^0=\text{MLP}_\text{encode}(\mathbf{x}_i^\text{input})
\end{equation}
In a second step, the processor of the GNS, which constitutes of $b$ message passing layers, iteratively updates the representation of the encoded node features $\mathbf{h}_i^r$, with $1\leq r\leq b$ in three steps:
\begin{enumerate}
    \item During the \textit{message passing} step the function $M$ with learnable parameters is used to compute a message vector $\mathbf{m}_{j,i}^{r+1}$ from each neighbour $v_j\in\mathcal{N}(v_i)$ to node $v_i$ by using the information of $\mathbf{h}_i^r$, $\mathbf{h}_j^r$, $\mathbf{e}_{j,i}$:
    \begin{equation}\label{eq:messagePassingStep}
    \mathbf{m}_{j,i}^{r+1}=M(\mathbf{h}_j^r,\mathbf{h}_i^r,\mathbf{e}_{j,i})        
    \end{equation}
    \item In the \textit{aggregation} step all messages are collected using a permutation invariant function $\oplus$, e.g. the mean of all messages:
    \begin{equation}\label{eq:aggregationStep}
        \mathbf{m}_i^{r+1}=\underset{v_j\in\mathcal{N}(v_i)}{\bigoplus} \mathbf{m}_{j,i}^{r+1}
    \end{equation}
    \item An \textit{update} of the node feature vector is generated using another learnable function $U$:
    \begin{equation}\label{eq:updateStep}
        \mathbf{h}_i^{r+1}=U(\mathbf{h}_i^{r},\mathbf{m}_i^{r+1})
    \end{equation}
\end{enumerate}
The final node representation $\mathbf{h}_i^b$ is decoded using $\text{MLP}_\text{decode}$:
\begin{equation}\label{eq:decoder}
    \mathbf{x}_i^\text{output}=\text{MLP}_\text{decode}(\mathbf{h}_i^b)
\end{equation}
Usually, the learnable functions $M$ and $U$ are also MLPs. Consequently, the structure of the GNS is given by the topology of these two MLPs, the encoder, the decoder, the aggregation function, and the number of message passing steps $b$. Message passing is a local exchange of information between connected neighbours. Consequently, a larger number of message passing steps allows information to propagate further across the graph. However, this might lead to over-smoothing \cite{cao2023efficient}.

So far, equations \ref{eq:encoder} to \ref{eq:decoder} describe a purely spatial processing of the graph data: $\mathbf{X}_\text{output}=\text{GNS}(\mathbf{A},\mathbf{X}_\text{input},\mathbf{E})$. To simulate a temporal evolution from time $t$ to $t+1$ the GNS model is applied to autoregressively predict the residual between two time steps:
\begin{equation}\label{eq:autoregression}
    \mathbf{X}_{t+1}=\mathbf{X}_t+\text{GNS}(\mathbf{A},\mathbf{X}_t,\mathbf{G},\mathbf{E})
\end{equation}
Note that, in this work we assume that only the node features are varying in time. However, it is possible to also use temporally dependent edge features. $\mathbf{G}\in\mathbb{R}^{n,g}$ is an additional matrix with $g$ temporally invariant (geometrical) node features.

While Equation \ref{eq:autoregression} describes an autonomous system (compare Equation \ref{eq:autonomousSystem}) where the node features only contain the flow state $\mathbf{X}_t=\mathbf{Y}_t$ we are interested in non-autonomous systems. Therefore, exogenous inputs can be incorporated into the structure by combining information about the flow field and the controls at a time $t$ into the node features: $\mathbf{x}_{i}^\text{input}=\text{concat}(\mathbf{y}_{i,t},\mathbf{u}_t)$ such that $\mathbf{x}_{i,t}\in\mathbb{R}^{q+c}$. In this work controls independent of the spatial dimension are used. If spatially varying controls, e.g. deforming geometries, are considered, the global control $\mathbf{u}_{t}$ would not be broadcasted to each node, but local control $\mathbf{u}_{i,t}$ could be used instead. While the information about the controls is used for message passing, the decoder only returns $\mathbf{y}_{i,t+1}-\mathbf{y}_{i,t}=\mathbf{x}_{i}^\text{output}$. Thus, the time stepping of the GNS with control inputs is given by:
\begin{equation}\label{eq:autoregressionWithControls}
    \mathbf{Y}_{t+1}=\mathbf{Y}_t+\text{GNS}(\mathbf{A},\mathbf{Y}_t,\mathbf{G},\mathbf{u}_{t},\mathbf{E})
\end{equation}

In contrast to steady ML models the temporal dimensions must be carefully considered for training. Analogous to \cite{pfaff2020learning} during training the parameters of the GNS are updated based on the next-step loss. This is teacher forcing \cite{lamb2016professor}, i.e. the model always uses the ground truth as an input. To improve rollout stability the inputs are perturbed with noise from a normal distribution $\mathcal{N}(0,\sigma^2_\text{noise})$ with the noise level $\sigma^2_\text{noise}$ which must be chosen \cite{sanchez2020learning}. Thus, the model is expected to learn how to handle small errors of the input during inference. Unrolled training would be an alternative option that supposedly leads to improved rollout stability \cite{kohl2026benchmarking}. Algorithm \ref{alg:GNSTraining} shows the GNS training. The model inference is not described separately, but is analogous to the autoregressive rollout during validation.

\begin{algorithm}
\caption{GNS Training (some aspects including data scaling and early stopping are not shown)}\label{alg:GNSTraining}
\begin{algorithmic}[1]
\ForAll{$\text{samples}_\text{non-uniform}$ in $\mathcal{D}$}
    \Comment{\textbf{pre-processing}}
    \State $\text{samples}_\text{uniform}\leftarrow\text{samples}_\text{non-uniform}$ \Comment{interpolate to uniform time grid}
\EndFor
\State $\mathcal{D}_{\text{train}},\mathcal{D}_{\text{validation}}\leftarrow\mathcal{D}$ \Comment{split dataset}
\ForAll{epochs}
    \ForAll{$\{\mathbf{Y}_0,\dots, \mathbf{Y}_T\}$, $\{\mathbf{u}_0,\dots,\mathbf{u}_T\}$ in $\mathcal{D}_{\text{train}}$}
    \Comment{\textbf{training (teacher forcing)}}
        \For{$t = 0$ \textbf{to} $T-1$}
            \State $\boldsymbol{\epsilon}\sim\mathcal{N}(0,\sigma^2_\text{noise})$ \Comment{compute noise}
            \State $\Delta\mathbf{\hat{Y}}_{t+1} \leftarrow \text{GNS}_\theta(\mathbf{A}, \mathbf{Y}_t+\epsilon,\mathbf{G}, \mathbf{u}_t, \mathbf{E})$
            \State  $\mathbf{\hat{Y}}_{t+1} \leftarrow \mathbf{Y}_t+\epsilon + \Delta\mathbf{\hat{Y}}_{t+1}$
        \EndFor
        \State $\mathcal{L}_\text{train} \leftarrow \text{lossFunction}(\{\mathbf{Y}_1,\dots, \mathbf{Y}_T\}, \{\mathbf{\hat{Y}}_1,\dots,\mathbf{\hat{Y}}_T\})$
        \State update weights $\theta$ via backpropagation and an optimiser
    \EndFor
    \ForAll{$\{\mathbf{Y}_0,\dots, \mathbf{Y}_T\}$, $\{\mathbf{u}_0,\dots,\mathbf{u}_T\}$ in $\mathcal{D}_{\text{validation}}$}
    \Comment{\textbf{validation (autoregressive)}}
        \State $\mathbf{\hat{Y}}_0\leftarrow\mathbf{Y}_0$\Comment{ initialise rollout prediction}
        \For{$t = 0$ \textbf{to} $T-1$}
            \State $\Delta\mathbf{\hat{Y}}_{t+1}\leftarrow\text{GNS}_\theta(\mathbf{A}, \mathbf{\hat{Y}}_t,\mathbf{G}, \mathbf{u}_t, \mathbf{E})$
            \State $\mathbf{\hat{Y}}_{t+1} \leftarrow \mathbf{\hat{Y}}_t + \Delta\mathbf{\hat{Y}}_{t+1}$
        \EndFor
        \State $\mathcal{L}_\text{validation} \leftarrow \text{validationLossFunction}(\{\mathbf{Y}_1,\dots, \mathbf{Y}_T\}, \{\mathbf{\hat{Y}}_1,\dots,\mathbf{\hat{Y}}_T\})$
    \EndFor
\EndFor
\end{algorithmic}
\end{algorithm}

\subsection{Neural Ordinary Differential Equations}\label{subsec:neuralOrdinaryDifferentialEquations}
Autoregressive \textit{next-step} models like the GNS (see Subsection \ref{subsec:graphNetworkSimulator}) are an Euler discretisation of a continuous problem: $\mathbf{y}_{t+1}=\mathbf{y}_t+hf_\theta(\mathbf{y}_t)$ with a fixed step size $h=1$ and a parametrisable function $f_\theta$, i.e. the GNS model in our case. However, temporal integration based on an Euler discretisation has a global error that is proportional to the step size. This can be overcome by taking the limit of the step size $h\rightarrow0$ such that the parametrised function learns the continuous dynamics instead of a stepwise evolution:
\begin{equation}\label{eq:continousDynamics}
    \frac{d\mathbf{y}(t)}{dt}=f_\theta(\mathbf{y}(t),t)
\end{equation}
In \cite{chen2018neural} this approach is introduced as Neural Ordinary Differential Equations (Neural ODEs), a model structure that allows for combining advantages of ML methods with the knowledge about ODE solvers:
\begin{align}\label{eq:neuralODE}
    \mathbf{y}(t_1)&=\mathbf{y}(t_0)+\int_{t_0}^{t_1}f_\theta(\mathbf{y}(t),t)dt \\
    &=\text{ODESolve}(\mathbf{y}(t_0),f_\theta,t_0,t_1)
\end{align}
Note that, $\text{ODESolve}$ might be any method to solve an ODE. This allows for the use of adaptive step sizes, stability guarantees and knowledge about error growth. Such a model can be trained by back-propagating through the solver or by using the $\mathcal{O}(1)$ memory efficient adjoint sensitivity method \cite{pontryagin1987mathematical} and afterwards continuously evaluated at any time $t$.

To expand the Neural ODE with control inputs the dynamics function must accept an additional input:
\begin{align}\label{eq:neuralODEWithControl}
    \frac{d\mathbf{y}(t)}{dt}=f_\theta(\mathbf{y}(t),\mathbf{u}(t),t)
\end{align}
Next-step models with control as in Equation \ref{eq:autoregressionWithControls} are evaluated at fixed times $t$. For such models, it is sufficient to have a discrete representation of control inputs. Depending on the ODE solver the dynamics function can be evaluated at any time $t$ which might not be represented in the ground truth dataset. Thus, an analytic or interpolated control function $f_u:t\rightarrow \mathbf{u}(t)$ that can be evaluated at any time is required:
\begin{align}\label{eq:neuralODEWithControls}
    \mathbf{y}(t_1)&=\text{ODESolve}(\mathbf{y}(t_0),f_\theta,f_u,t_0,t_1)
\end{align}

The Markov property states that if the entire state of a system is known at one time all future states are uniquely defined by the current state \cite{oksendal2003stochastic}. In case the system is only partly observed the Markov property is violated, i.e. modelling such an under-resolved system using a model as in Equation \ref{eq:continousDynamics} is not a \textit{well-posed} problem, because the solution is non-unique. This is true in case of a sub-selection of spatial points or if a sub-selection of physical quantities is considered. For instance, a single surface pressure snapshot may correspond to multiple distinct full-flow states during hysteresis loops. Introducing augmented latent dimensions is an approach to deal with the violated Markov property. According to literature, augmented Neural ODE models can be more expressive, have reduced computational effort and better generalisation than a standard Neural ODE \cite{dupont2019augmented}. Instead of just evolving the state $\mathbf{y}(t)\in\mathbb{R}^{q}$ through time, latent variables $\mathbf{l}(t)\in\mathbb{R}^l$ are used to create a lifted state $\left[\mathbf{y}(t),\mathbf{l}(t)\right]^\top\in\mathbb{R}^{q+l}$:
\begin{equation}\label{eq:anodeDynamics}
    \frac{d}{dt}\begin{bmatrix}\mathbf{y}(t)\\ \mathbf{l}(t)\end{bmatrix} = f_\theta\left(\begin{bmatrix}\mathbf{y}(t)\\ \mathbf{l}(t)\end{bmatrix},\mathbf{u}(t),t\right).
\end{equation}
The augmented dimensions $\mathbf{l}(t_0)$ are usually initialised to zero. During training the augmented Neural ODE learns to utilise the augmented dimensions. In training the parameters $\theta$ are updated based on a loss $\mathcal{L}$, usually the reconstruction loss, e.g. mean squared error between ground truth and prediction. To account for the augmented states in the loss function a regularisation term weighted by $\lambda$ is introduced:
\begin{align}\label{eq:augmentedLoss}
    \mathcal{L}&=\mathcal{L}_\text{Reconstruction}(\mathbf{y}_\text{truth},\mathbf{y}_\text{predicted})+\lambda\mathcal{L}_\text{Regularisation}(\mathbf{l})\\
    &=\text{MSE}(\mathbf{y}_\text{truth},\mathbf{y}_\text{predicted})+\lambda\|\mathbf{l}\|_2
\end{align}

\subsection{Graph Neural ODE}\label{subsec:graphNeuralODEs}

\begin{figure}
    \centering
    \includegraphics[width=0.95\textwidth]{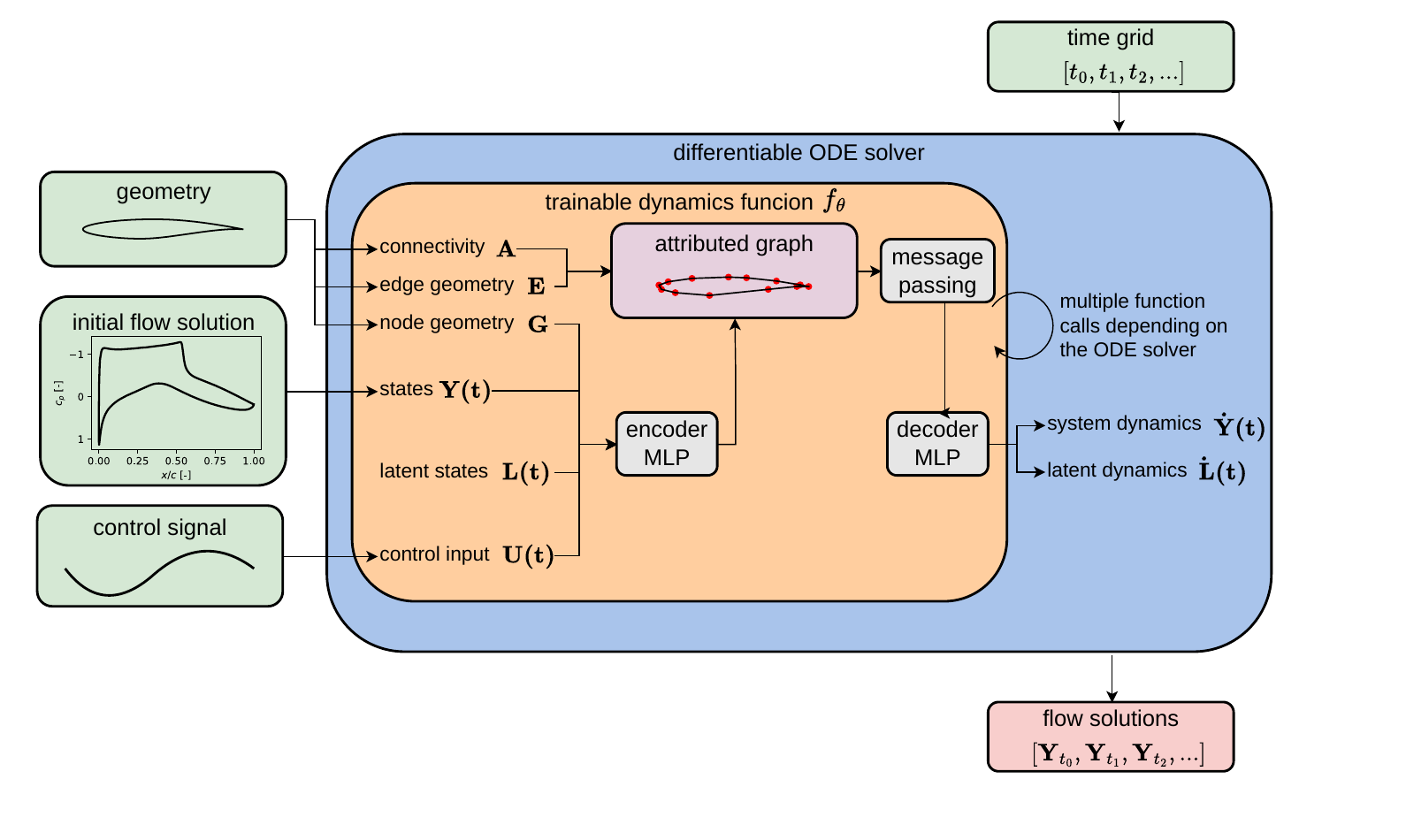}
    \caption{Schematic overview of the GNODE approach with augmented dimensions (see Equation \ref{eq:dynamicsGNODE}). Note that the inner structure of the ODE solver is not depicted.}
    \label{fig:methodGNODE}
\end{figure}

The GNS (see Subsection \ref{subsec:graphNetworkSimulator}) uses message passing to perform \textit{spatial} processing of data. However, the time-stepping scheme is basic. Neural ODEs (with controls and augmented dimensions) provide an advanced method for \textit{temporal} processing of data. This work investigates a combination of both approaches to obtain a strong framework for \textit{spatio-temporal} data processing, called Graph Neural ODE (GNODE).
A GNS structured model with control inputs (see Equation \ref{eq:autoregressionWithControls}) is used as the parametrisable function $f_\theta$ in Equation \ref{eq:neuralODEWithControl}, respectively in Equation \ref{eq:anodeDynamics} with augmented dimensions:
\begin{equation}\label{eq:dynamicsGNODE}
    \frac{d}{dt}\begin{bmatrix}\mathbf{Y}(t)\\ \mathbf{L}(t)\end{bmatrix}=\text{GNS}_\theta\left(\mathbf{A},\begin{bmatrix}\mathbf{Y}(t)\\ \mathbf{L}(t)\end{bmatrix},\mathbf{G},\mathbf{u}(t),\mathbf{E}\right)
\end{equation}

The ODE solver evolves $\begin{bmatrix}\mathbf{Y}(t)\\ \mathbf{L}(t)\end{bmatrix}\in\mathbb{R}^{n,q+l}$ in time. Figure \ref{fig:methodGNODE} visualises the GNODE approach. Algorithm \ref{alg:GNODETraining} presents the training of the GNODE approach. Inference is analogous to the validation loop. For each function evaluation of the GNS the node features are constructed from the physical state $\mathbf{y}(t)\in\mathbb{R}^q$, the augmented latent dimensions $\mathbf{l}(t)\in\mathbb{R}^l$, static geometrical features $\mathbf{g}\in\mathbb{R}^g$, and control inputs $\mathbf{u}\in\mathbb{R}^c$: $\mathbf{x}_i^\text{input}(t)\in\mathbb{R}^{q+l+g+c}$. After passing the data through the GNS the output contains the derivatives $\frac{d\mathbf{y}(t)}{dt}\in\mathbb{R}^q$ and $\frac{d\mathbf{l}(t)}{dt}\in\mathbb{R}^l$: $\mathbf{x}_i^\text{output}(t)\in\mathbb{R}^{q+l}$.

\begin{algorithm}
\caption{GNODE Training (some aspects including data scaling and early stopping are not shown)}\label{alg:GNODETraining}
\begin{algorithmic}[1]
\State $\mathcal{D}_{\text{train}},\mathcal{D}_{\text{validation}}\leftarrow\mathcal{D}$ \Comment{split dataset}
\Function{analyticControl}{$t$, $\text{context}$} \Comment{$\text{context}$ may contain $\alpha$, $\hat{\alpha}$, $k$}
    \State $\dots$
    \State \Return $\mathbf{u}_t$
\EndFunction
\Function{dynamicsFunction}{$t$, [$\mathbf{Y}_t$, $\mathbf{L}_t$], $\text{context}$} \Comment{called by $\text{ODESolve}$}
    \State $\mathbf{u}_t \leftarrow \text{analyticControl}(t,\text{context})$
    \State $[\frac{d\mathbf{Y}_t}{dt},\frac{d\mathbf{L}_t}{dt}]=\text{GNS}_\theta\left(\mathbf{A},[\mathbf{Y}_t,\mathbf{L}_t],\mathbf{G},\mathbf{u}_t,\mathbf{E}\right)$
    \State \Return $[\frac{d\mathbf{Y}_t}{dt},\frac{d\mathbf{L}_t}{dt}]$
\EndFunction
\ForAll{epochs}
    \ForAll{$\{\mathbf{Y}_0,\dots, \mathbf{Y}_T\}$, $\text{context}$, $\{t_0,\dots, t_T\}$ in  $\mathcal{D}_{\text{train}}$} \Comment{\textbf{training}}
        \State $\mathbf{L}_0\leftarrow\mathbf{0}$ \Comment{initialise latent states}
        \State $\begin{aligned}[t][\{\mathbf{\hat{Y}}_0,\dots, \mathbf{\hat{Y}}_T\},\{\mathbf{L}_0,\dots, \mathbf{L}_T\}] \leftarrow\text{ODESolve}(& \text{dynamicsFunction}(\text{context}=\text{context}),\\& [\mathbf{Y}_0, \mathbf{L}_0], \{t_0,\dots, t_T\})\end{aligned}$
        \State $\mathcal{L}_\text{train} \leftarrow \text{lossFunction}(\{\mathbf{Y}_1,\dots, \mathbf{Y}_T\}, \{\mathbf{\hat{Y}}_1,\dots,\mathbf{\hat{Y}}_T\},\{\mathbf{L}_1,\dots,\mathbf{L}_T\})$
        \State update weights $\theta$ via backpropagation through $\text{ODESolve}$ and an optimiser
    \EndFor
    \ForAll{$\{\mathbf{Y}_0,\dots, \mathbf{Y}_T\}$, $\text{context}$, $\{t_0,\dots, t_T\}$ in $\mathcal{D}_{\text{validation}}$} \Comment{\textbf{validation}}
        \State $\mathbf{L}_0\leftarrow\mathbf{0}$ \Comment{initialise latent states}
        \State $\begin{aligned}[t][\{\mathbf{\hat{Y}}_0,\dots, \mathbf{\hat{Y}}_T\},\{\mathbf{L}_0,\dots, \mathbf{L}_T\}] \leftarrow\text{ODESolve}(& \text{dynamicsFunction}(\text{context}=\text{context}),\\& [\mathbf{Y}_0, \mathbf{L}_0], \{t_0,\dots, t_T\})\end{aligned}$
        \State $\mathcal{L}_\text{validation} \leftarrow \text{validationLossFunction}(\{\mathbf{Y}_1,\dots, \mathbf{Y}_T\}, \{\mathbf{\hat{Y}}_1,\dots,\mathbf{\hat{Y}}_T\})$
    \EndFor
\EndFor
\end{algorithmic}
\end{algorithm}

Table \ref{tab:methodComparison} outlines structural differences between the GNODE approach (see Subsection \ref{subsec:graphNeuralODEs}) and the autoregressive GNS (see Subsection \ref{subsec:graphNetworkSimulator}). Both models are implemented using the following \texttt{python} packages: \texttt{PyTorch} \cite{paszke2019pytorch}, \texttt{PyTorch Geometric} \cite{fey2019fast}, \texttt{torchdiffeq} \cite{chen2018torchdiffeq}. Furthermore, the Surrogate Modeling for AeRo data Toolbox Python Package (\texttt{SMARTy}) is used \cite{bekemeyer2022data, bekemeyer2024recent}.

\begin{table}
    \centering
    \setlength{\leftmargini}{0.4cm}
    \begin{tabular}{m{0.15\textwidth}|m{0.35\textwidth}|m{0.35\textwidth}}
          & GNS & GNODE \\ \hline
          time step size & 
          \begin{itemize}
              \item The model learns to predict $\Delta\mathbf{Y}_t$ for a fixed $\Delta t$.
              \item Training data must have a uniform time grid $\Delta t=\text{constant}$
              \item If the training dataset contains variable time step sizes interpolation is required.
              \item During inference the GNS model is limited to the time step size it was trained on.
          \end{itemize} & 
          \begin{itemize}
              \item The approach is not limited to a fixed time step size since the derivative is predicted.
              \item For batch training all samples within a batch must have the same time grid. In this work, each training batch does only contain one sample.
          \end{itemize} \\ \hline
          control signal & 
          \begin{itemize}
              \item Discrete control inputs $u_t$ at each $t$ which is present in the training data can be used. 
              \item If temporal interpolation due to different time step sizes is necessary, this also affects the control inputs.
          \end{itemize} & 
          \begin{itemize}
              \item Depending on the chosen method for $\text{ODESolve}$ the model $f_\theta$ (see Equation \ref{eq:anodeDynamics}) may be evaluated at any time $t$. Thus, a functional representation of the control inputs $\mathbf{u}(t)$ is required.
              \item If known $\mathbf{u}(t)$ can be an analytical representation. Else it may also be an interpolated function of discrete controls from the training data.
          \end{itemize} \\ \hline
          training methodology & 
          \begin{itemize}
              \item The model is trained using the teacher forcing methodology. Thus, only the next step error is taken into account.
              \item Noise augmentation is required and introduces an additional hyperparameter.
          \end{itemize} & 
          \begin{itemize}
              \item Errors are propagated through the solver using backpropagation or the adjoint method.
          \end{itemize} \\ \hline
          augmented dimensions & 
          \begin{itemize}
              \item No augmented dimensions are used for this approach.
          \end{itemize} & 
          \begin{itemize}
              \item Augmented dimensions are used for this approach.
          \end{itemize} \\
    \end{tabular}
    \caption{Comparison of selected differences between GNS and GNODE.}
    \label{tab:methodComparison}
\end{table}

\section{Data}\label{sec:data}
\begin{figure}
    \centering
    \includegraphics[width=0.45\textwidth]{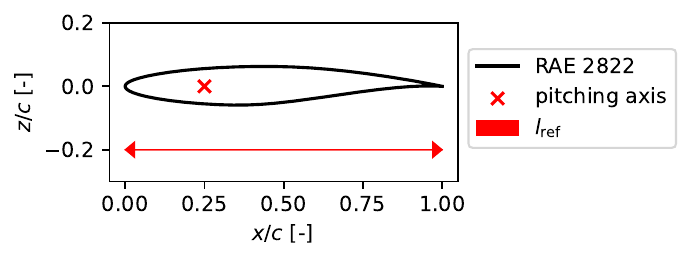}
    \caption{RAE 2822 airfoil. The pitch axis is located at $x/c=0.25,z/c=0$. The reference length is $l_\text{ref}=c=1\,\text{m}$.}
    \label{fig:rae2822Airfoil}
\end{figure}

In addition to model architecture, data-driven methods require meaningful training data to achieve good results. For this study, a dataset is created containing two-dimensional, unsteady simulations of the RAE 2822 airfoil. The rear-loaded, sub-critical RAE 2822 airfoil has a rooftop-type pressure distribution at its steady design conditions $M_{\infty,\text{Design}}=0.66$, $c_{l,\text{Design}}=0.56$ \cite{cook1977aerofoil}. This airfoil is a common test case for numerical methods as experimental reference data is available. The CFD simulations in this study are conducted using the medium grid with $n_\text{surface}=512$ grid points on the airfoil surface from \cite{langer2019initial}. The unstructured mesh is chosen based on a mesh-convergence study that compares the results of steady simulations on different meshes at design conditions to experimental data. Figure \ref{fig:rae2822Airfoil} visualises the airfoil.

\begin{figure}
    \centering
    \includegraphics[width=0.55\textwidth]{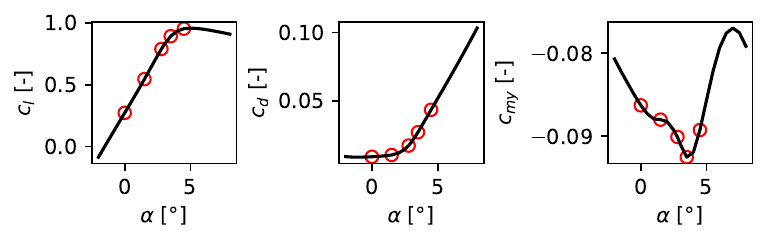}
    \caption{Resulting global coefficients ($c_l$, $c_d$, $c_{my}$) of steady simulations. The red circles show the results at selected initial angles of attack.}
    \label{fig:steadyPolars}
\end{figure}

It is of interest to find a surrogate model that delivers predictions at various (unsteady) flight conditions. Therefore, the dataset which is used to train the model must reflect the complexity of the target application. This work uses a dataset consisting of unsteady forced-motion aerodynamic simulations, each starting from a converged steady simulation with different angles of attack $\alpha_0$.
Unsteadiness is caused by a prescribed sinusoidal pitching motion $\alpha(t)$. The pitching motion occurs about the quarter-chord point of the airfoil and is parametrised by the amplitude $\hat{\alpha} $ and the circular frequency $\omega$:
\begin{equation}\label{eq:pitchingMotion}
    \alpha(t)=\alpha_0+\hat{\alpha}\sin{(\omega t)}
\end{equation}
The reduced frequency $k$ is a dimensionless number that is commonly used during the analysis of unsteady aerodynamics:
\begin{equation}\label{eq:reducedFrequency}
    k=\frac{\omega l_\text{ref}}{u_\text{ref}}
\end{equation}
where $l_\text{ref}$ is a reference length, here the chord length, and $u_\text{ref}$ is the reference airspeed.
A three-dimensional full-factorial DoE is created with the following variables:
\begin{itemize}
    \item initial steady angle of attack: $\alpha_0\in\{0.0^\circ,1.5^\circ,2.79^\circ, 3.5^\circ,4.5^\circ\}$.
    \item reduced frequency:  $k\in\{0.1,0.2,0.3,0.5,1.0\}$.
    \item amplitude: $\hat{\alpha}\in\{0.1^\circ,1.0^\circ,2.0^\circ\}$.
\end{itemize}
This results in a dataset with $5\cdot5\cdot3=75$ unique unsteady trajectories. The reference data is produced using the DLR-TAU code \cite{schwamborn2008development} which solves the (U)RANS-equations using the negative Spalart-Allmaras turbulence model \cite{allmaras2012modifications}. Before unsteady simulations are computed, first five steady simulations are conducted at transonic flight conditions ($M_\infty=0.73$, $Re=6.5\cdot10^6$). These conditions align with the experiments from \cite{cook1977aerofoil} and Figure \ref{fig:steadyPolars} presents the resulting global coefficients. The five steady simulations are located at different flight conditions. While $\alpha_0=0.0^\circ$ and $\alpha_0=1.5^\circ$ are within the linear range of the lift curve, $\alpha_0=3.5^\circ$ and $\alpha_0=4.5^\circ$ are close to $c_{l,\max}$. Similarly, the middle plot shows that the simulations also cover a range of different physical behaviour regarding the coefficient of drag. The most complex behaviour can be seen for the pitching moment coefficient. Up to $\alpha=3.5^\circ$  $c_{my}$ decreases; afterwards, it increases. Consequently, as evident by the global coefficients, already the steady simulations of the dataset cover a range of physical effects.

Starting from these converged steady simulations, subsequently, forced-motion unsteady flow solutions are computed using a dual-time stepping scheme \cite{jameson2015application, yang1995implementation}. The solution is stepped forward in time using the second-order backwards differentiation formula (BDF2) which is an A-stable implicit integration scheme \cite{curtiss1952integration}. Thus, the time step size $\Delta t$ is only limited by accuracy considerations, not by stability considerations. Details regarding the CFD simulations are presented in the Appendix \ref{subsec:CFDSimulationDetails}. To establish a baseline for computational efficiency, the cost of generating the reference data is quantified. The simulation of a single physical time step required an average of 145\,s on an AMD EPYC 9555 CPU. Consequently, the average total runtime per unsteady trajectory was 20\,h and 43\,min. 

\begin{figure}
    \centering
    \includegraphics[width=0.35\textwidth]{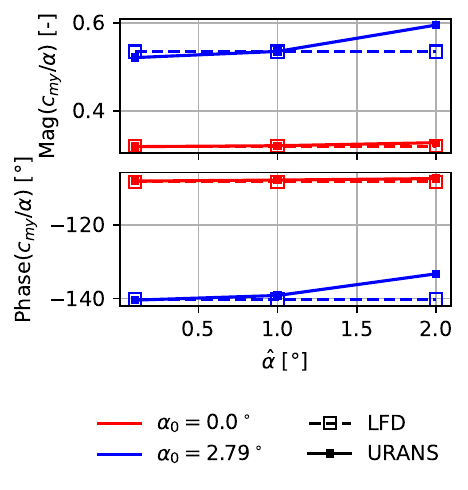}
    \caption{Comparison of the first harmonic of a Fourier transformation of URANS simulation versus LFD solution at different amplitudes.}
    \label{fig:URANSvsLFD}
\end{figure}

Figure \ref{fig:URANSvsLFD} compares a subset of the URANS results at different amplitudes with results obtained using the LFD solver (see Section \ref{sec:introduction} and \cite{thormann2013linear} for further details) which is only valid for small amplitudes. For $\alpha_0=0^\circ$ LFD and URANS yield the same answer. Thus, the part of the solution after the transients have decayed shows a dynamically linear behaviour. Conversely, at $\alpha_0=2.79^\circ$ and $\hat{\alpha}=2.0^\circ$ LFD and URANS do not align. Consequently, the URANS result shows dynamically non-linear behaviour. This underlines the motivation to find a surrogate that is capable of predicting solutions in both, the dynamically linear (small amplitudes) and non-linear region (large amplitudes).

The present study utilises a restricted subset of the full dataset to focus the analysis on specific dynamic regimes. First, the samples at $\alpha_0=4.5^\circ$ are excluded as these samples are very close to $c_{l,\max}$ and thus show a very different physical behaviour than the other samples, i.e. separation. Furthermore, $\hat{\alpha}=0.1^\circ$ is omitted. According to Figure \ref{fig:URANSvsLFD} dynamically linear and non-linear behaviour is still covered with the remaining two amplitudes. The remaining $4\cdot5\cdot2=40$ samples are split into a training and test sets. One reduced frequency is used for testing: $k_\text{test}=0.3$ ($4\cdot1\cdot2=8$ test samples).

\section{Results}\label{sec:results}
This Section evaluates the proposed GNODE framework against the GNS baseline using the RAE 2822 dataset described in Section \ref{sec:data}. GNS is selected as a baseline because its discrete-time formulation serves as a natural counterpart to the continuous-time dynamics of the GNODE, allowing for an assessment of the temporal integration method.

\subsection{Model Selection}\label{subsec:modelSelection}
The model topology, e.g. the number of layers and neurons, and the training dynamics are influenced by numerous hyperparameters. A hyperparameter optimisation (HPO) is performed for both the GNS and the GNODE framework to allow for a fair comparison by using a suitable set of hyperparameters. Therefore, each architecture is trained 160 times on the training dataset for a maximum of 1000 epochs per trial. The training is stopped if the validation metric, i.e. the mean absolute error (MAE) on the validation data, does not improve for 100 epochs (early stopping). Table \ref{tab:modelSelection} outlines the search spaces and the resulting best hyperparameters. The optimisation framework \texttt{optuna} is used \cite{akiba2019optuna}.

\begin{table}
    \centering
    \begin{tabular}{c|c|c|c|c|c}
         \textbf{hyperparameter} & \textbf{GNS} & \textbf{GNODE} & \textbf{possible values} & \textbf{result GNS} & \textbf{result GNODE} \\ \hline
         $\text{augmentedDimension}$ &  & $\checkmark$ & integer $\in[0,20]$ &  & $14$ \\
         $\text{latentDimension}$ & $\checkmark$ & $\checkmark$ & integer $\in[16,256]$ & $203$ & $186$ \\
         $\text{numberNeurons}$ & $\checkmark$ & $\checkmark$ & integer $\in[16,256]$ & $256$ & $86$ \\
         $\text{numberLayersProcessor}$ & $\checkmark$ & $\checkmark$ & integer $\in[1,3]$ & $2$ & $1$ \\
         $\text{numberHiddenLayersEnDeCoder}$ & $\checkmark$ & $\checkmark$ & integer $\in[0,1]$ & $1$ & $0$ \\
         $\text{activation}$ & $\checkmark$ & $\checkmark$ & $[\text{tanh},\text{relu}]$ & $\text{relu}$ & $\text{tanh}$ \\
         $\text{regularisationFactor}$ &  & $\checkmark$ & float $\in[1\cdot10^{-8}, 1\cdot10^{-3}]$ &  & $2.23\cdot10^{-6}$ \\
         $\text{optimiser}$ & $\checkmark$ & $\checkmark$ & $[\text{adam},\text{adamw}, \text{adamax}]$ & $\text{adamax}$ & $\text{adamax}$ \\
         $\text{learningRate}$ & $\checkmark$ & $\checkmark$ & float $\in[5\cdot10^{-5}, 5\cdot10^{-2}]$ & $9.44\cdot10^{-4}$ & $1.27\cdot10^{-2}$ \\
         $\text{weightDecay}$ & $\checkmark$ & $\checkmark$ & float $\in[1\cdot10^{-8}, 1\cdot10^{-3}]$ & $6.93\cdot10^{-6}$ & $9.91\cdot10^{-6}$ \\
         $\text{gamma}$ & $\checkmark$ & $\checkmark$ & float $\in[0.9, 0.9999]$ & $0.9415$ & $0.991$ \\
         $\text{noiseLevel}$ & $\checkmark$ &  & float $\in[0.0, 0.1]$ & $0.0183$ & \\
    \end{tabular}
    \caption{Model selection for the GNS and GNODE method.}
    \label{tab:modelSelection}
\end{table}

To verify the outcome of the HPO the resulting values for selected hyperparameters are discussed. Firstly, the resulting activation function for the GNS is the rectified linear unit (ReLU \cite{agarap2018deep}) which is in accordance with the established literature \cite{pfaff2020learning}. Conversely, the continuous-time GNODE uses a hyperbolic tangent (tanh) as activation function. This choice is expected for continuous-time dynamical modelling where the smoothness of the activation function is a critical property for stable training dynamics of Neural ODEs \cite{gao2025global}. Secondly, for both methodologies the HPO identifies Adamax as the best optimiser. Compared to standard Adam, Adamax is more robust to vanishing and exploding gradients by utilising the infinity-norm \cite{kingma2014adam}. This is crucial for the Neural ODE-based approach where gradients are propagated through the continuous-time ODE solver. Finally, the resulting GNS architecture exhibits multiple hidden layers within the encoder, processor and decoder. This results in 735226 trainable network parameters. In contrast, the best GNODE model as identified by the HPO is the shallowest possible configuration, i.e. the model has 104507 trainable parameters. The GNN inside the GNODE parametrises the dynamics. Deep GNNs can produce vector fields with high non-linearity, which may lead to numerical stiffness and instability in the ODE solver. Stiffness leads to a more complex optimisation problem during training. Additionally, the GNODE is augmented with 14 latent dimensions. The effect of this parameter is further explored in Subsection \ref{subsec:augmentedDimensions}.

\subsection{Full Dataset Test}\label{subsec:fullDatasetTest}
Using the optimal set of hyperparameters, as discussed above, the GNS and GNODE models are trained on the full training dataset (see Section \ref{sec:data}) for a maximum of 2000 epochs. Early stopping is applied to conserve computational resources, stopping training after 200 epochs without validation improvement. Furthermore, to account for variability introduced by random parameter initialisation and data splitting, five independent instances of each architecture are trained. The NVIDIA RTX 2000 Ada Generation GPU is used to train the GNS and GNODE models. Table \ref{tab:computationalEffort} presents the training and inference times. GNODE requires 5.3 times more time per epoch than GNS. GNODE employs a fourth-order Runge-Kutta (RK4) scheme, evaluating the internal GNN four times per integration step, whereas the autoregressive GNS is called once per physical time step. Additionally, GNS uses far less training epochs to reach convergence. Note that, all implementations are not optimised for optimal inference or training speed; however, they clearly illustrate the computational trade-offs between discrete and continuous-time architectures.

\begin{table}
    \centering
    \small
    \begin{tabular}{llcc}
        \toprule
        & & \textbf{GNS} & \textbf{GNODE} \\ 
        \midrule
        \multirow{3}{*}{\makecell[l]{\textbf{training}}}
        & time per epoch [s] & $5.4$ &  $28.5$ \\
        & epochs [-] & $237 \pm 14$ & $1550 \pm 348$ \\
        & runtime [h] & $\mathbf{0.36\pm0.02}$ & $9.11 \pm 2.76$ \\
        \midrule
        \multirow{1}{*}{\makecell[l]{\textbf{inference}}}
        & time per trajectory [s] & $0.81\pm0.19$ & $\mathbf{0.70 \pm 0.22}$ \\
        \bottomrule
    \end{tabular}
    \caption{Comparison of the computational effort for training and inference of GNS and GNODE models. \textbf{Bold} values are the best in a category. Mean ± standard deviation across five independent training runs.}
    \label{tab:computationalEffort}
\end{table}

\begin{figure}
\centering
\begin{subfigure}{0.32\textwidth}
    \includegraphics[width=\textwidth]{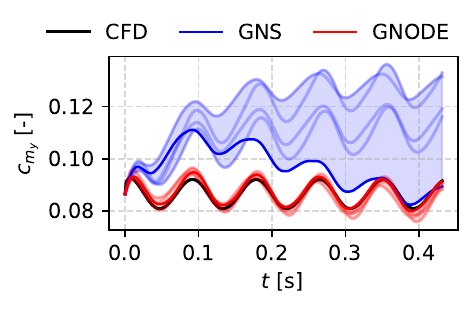}
    \caption{$\alpha_0=0.0^\circ$, $\hat\alpha=1.0^\circ$.}
    \label{subfig:coeff2Alpha0.0Amp1.0}
\end{subfigure}
\begin{subfigure}{0.32\textwidth}
    \includegraphics[width=\textwidth]{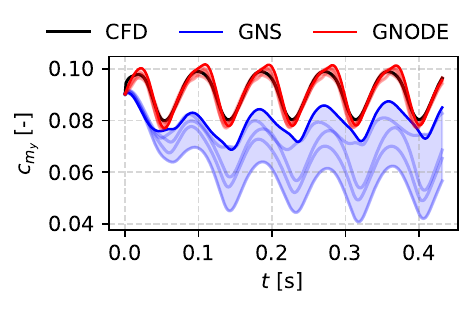}
    \caption{$\alpha_0=2.79^\circ$, $\hat\alpha=1.0^\circ$.}
    \label{subfig:coeff2Alpha2.79Amp1.0}
\end{subfigure}
\begin{subfigure}{0.32\textwidth}
    \includegraphics[width=\textwidth]{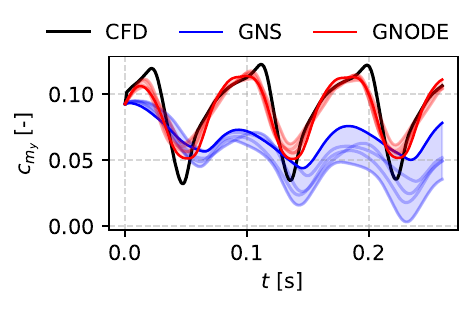}
    \caption{$\alpha_0=3.5^\circ$, $\hat\alpha=2.0^\circ$.}
    \label{subfig:coeff2Alpha3.5Amp2.0}
\end{subfigure}
\caption{Pitching moment coefficient $c_{my}$ over time $t$. $c_{my}$ is calculated using $c_p$, $c_{f,x}$ and $c_{f,z}$. For GNS and GNODE the dark-coloured lines correspond to the best of five models according to the metrics in table \ref{tab:metricsFullDatasetTest}. The light-coloured lines show the remaining four models. The shaded regions indicate the spread of predictions across the model instances.}
\label{fig:globalCoefficients}
\end{figure}

The trained models are evaluated using multiple metrics. Each metric in Table \ref{tab:metricsFullDatasetTest} is an average over the prediction of one model instance on the entire test dataset. Mean and standard deviation are calculated based on the resulting values of five model instances. Definitions of the metrics are provided in the Appendix \ref{subsec:metrics}. GNODE clearly outperforms GNS in all metrics. In addition to superior mean values GNODE shows lower standard deviations. This suggests that GNODE exhibits lower prediction variance and more consistent performance across random initialisations. Notably, the coefficient of determination for the global quantities of the GNS approach yields a negative value. As visible in Figure \ref{fig:globalCoefficients} the integrated global coefficients of the GNS predictions drift from the reference solution, which leads to unexpected values of the $R^2$-metric.

\begin{table}
    \centering
    \small
    \begin{tabular}{llcc}
        \toprule
        & & \textbf{GNS} & \textbf{GNODE} \\ 
        \midrule
        \multirow{3}{*}{\makecell[l]{\textbf{surface quantities} \\ \textit{averaged over} $c_{p}$, $c_{fx}$, $c_{fz}$}} 
        & MAE [$\cdot 10^{-3}$] $\downarrow$ & $19.75 \pm 2.40$ & $\mathbf{5.28 \pm 0.33}$ \\
        & MSE [$\cdot 10^{-4}$] $\downarrow$ & $32.14 \pm 7.67$ & $\mathbf{2.84 \pm 0.25}$ \\
        & $R^2$ [-] $\uparrow$ & $0.9721 \pm 0.0067$ & $\mathbf{0.9975 \pm 0.0002}$ \\
        \midrule
        \multirow{6}{*}{\makecell[l]{\textbf{global quantities} \\ \textit{evaluated on} $c_{my}$}}
        & MAE [$\cdot 10^{-3}$] $\downarrow$ & $21.81 \pm 5.04$ & $\mathbf{2.61 \pm 0.14}$ \\
        & MSE [$\cdot 10^{-5}$] $\downarrow$ & $69.62 \pm 27.13$ & $\mathbf{1.89 \pm 0.29}$ \\
        & $R^2$ [-] $\uparrow$ &$ -3.5118 \pm 1.7579$ & $\mathbf{0.8879 \pm 0.0184}$ \\
        & drift [$\cdot 10^{-3}$ s$^{-1}$] $\downarrow$ & $28.44 \pm 16.49$ & $\mathbf{5.39 \pm 2.83}$ \\
        & amplitude deviation [$\cdot 10^{-2}$] $\downarrow$ & $14.49 \pm 9.91$ & $\mathbf{5.48 \pm 3.23}$ \\
        & phase error [$^\circ$] $\downarrow$ & $22.22 \pm 8.86$ & $\mathbf{7.88 \pm 5.67}$ \\ 
        \bottomrule
    \end{tabular}
    \caption{Comparison of selected metrics between GNS and GNODE. Each model is trained five times. The metrics show mean $\pm$ standard deviation. \textbf{Bold} values present the best result for each metric. The arrows indicate whether low ($\downarrow$) or a high ($\uparrow$) values is desirable.}
    \label{tab:metricsFullDatasetTest}
\end{table}

\begin{figure}
    \centering
    \includegraphics[width=0.5\textwidth]{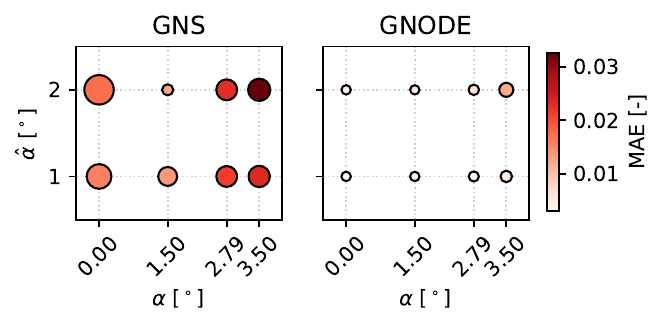}
    \caption{MAE of predicted values relative to CFD reference data averaged over the surface quantities ($c_p$, $c_{f,x}$, $c_{f,z}$). The plot shows the results of five model instances of GNS and GNODE. The colour scale corresponds to the mean values. The size of the circles corresponds to the standard deviation.}
    \label{fig:alphaAmpMAE}
\end{figure}

To evaluate the influence of input parameters $\alpha$ and $\hat{\alpha}$ on the prediction accuracy of the model, the MAE averaged over all surface quantities (as in Table \ref{tab:metricsFullDatasetTest}) is presented in Figure \ref{fig:alphaAmpMAE}. The overall magnitude of errors and the corresponding variance are roughly four times higher for GNS than for GNODE across the entire parameter space. Both models produce the highest errors at $\alpha_0=3.5^\circ$, $\hat{\alpha}=2.0^\circ$. This transonic flight condition is close to $c_{l,\max}$ and near $c_{my,\min}$. Combined with the high amplitude this represents the most challenging test case, i.e. the airfoil traverses different regimes of the steady lift and pitching moment curves and exhibits strong dynamic non-linearities. Consequently, GNS and GNODE generally show higher errors towards the higher angles of attack and amplitudes. The variations between multiple model instances of the GNS predictions does not exhibit a clear trend. However, for GNODE the variance in results is slightly larger for the more challenging cases. This demonstrates the superiority of GNODE over GNS.

While both models output surface quantities directly, global aerodynamic parameters are often the primary interest of aerospace engineers. Furthermore, scalar quantities are easier to interpret and visualise compared to field quantities and hence discussed next. Global quantities ($c_l$, $c_d$ and $c_{my}$) are obtained by integrating the surface pressure and skin friction coefficients over the airfoil surface. The pitching moment coefficient is particularly sensitive to local variations; therefore, it is selected here as the global metric for further analysis. Figure \ref{fig:globalCoefficients} presents the evolution of $c_{my}$ over time. Consistent with the prior results, the GNODE produces stable roll-outs with small variance throughout. Conversely, the GNS predictions diverge from the CFD reference already during the first oscillation period and never reaches a stable period behaviour. While the CFD reference results in Subfigures \ref{subfig:coeff2Alpha0.0Amp1.0} and \ref{subfig:coeff2Alpha2.79Amp1.0} follow the sinusoidal forcing input, the case shown in Subfigure \ref{subfig:coeff2Alpha3.5Amp2.0} does not exhibit purely sinusoidal behaviour. Consistent with the previous results the GNODE result is unable to fully capture the behaviour in this case. Nevertheless, despite some reduction in accuracy in this challenging case, the GNODE predictions remain temporally bounded and stable. Unlike GNODE, the GNS model is less robust to random initialisation, as evidenced by the high variance in its results. Additional results for more input parameter combinations are presented in the Appendix \ref{subsec:additionalResults}.

\begin{figure}
    \centering
    \includegraphics[width=0.75\textwidth]{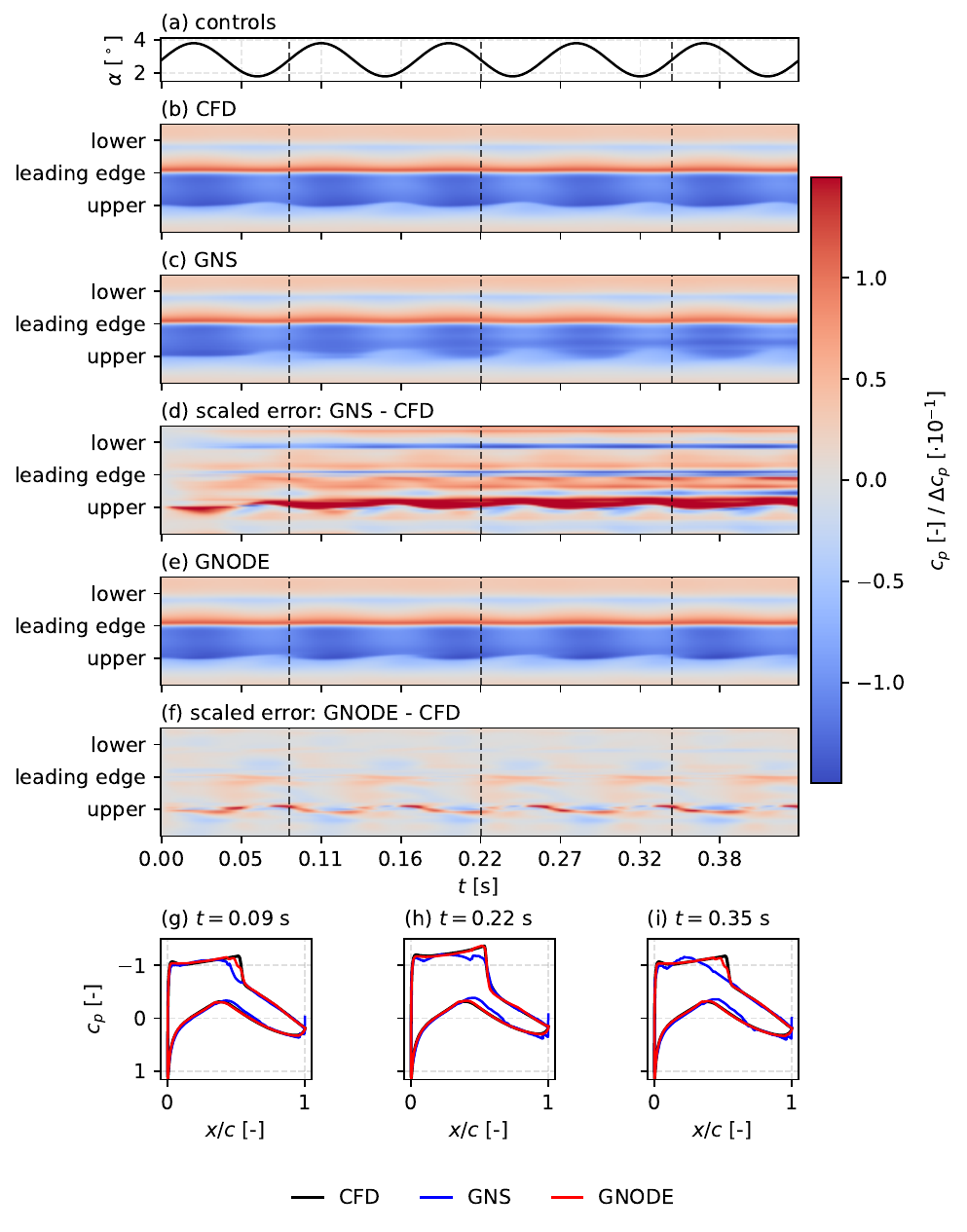}
    \caption{Spatio-temporal evolution of the surface quantities for $\alpha_0=2.79^\circ$, $\hat{\alpha}=1.0^\circ$. Panel (a) shows $\alpha$, which is part of the forcing control signal. Subfigures (b)--(f) show the evolution of the surface pressure over time: the y-axis corresponds to the \textit{unrolled} surface of the airfoil (compare Figure \ref{fig:rae2822Airfoil}). Panel (b) shows the CFD reference, while (c) and (e) show the predictions of the best GNS and GNODE models according to the metrics in Table \ref{tab:metricsFullDatasetTest}. Panels (d) and (f) show the error relative to the CFD reference solution. Subfigures (g), (h), and (i) present $c_p$-distributions along the length of the airfoil ($x/c$) at three times $t$. To highlight the errors, the values (and colours) are amplified by a factor of 10. The vertical dashed lines in (a)--(f) present the times $t$ corresponding to Subfigures (g)--(i).}
    \label{fig:surfacePlotAlpha2.79Amp1.0}
\end{figure}

After evaluating the models on several error metrics and global coefficients, the surface output is analysed in more depth. Figure \ref{fig:surfacePlotAlpha2.79Amp1.0} compares the surface pressure predictions of the two approaches over time at $\alpha_0=2.79^\circ$, $\hat{\alpha}=1.0^\circ$. Subfigures (b)–(f) show the evolution of the surface pressure over time: The y-axis corresponds to the unrolled surface of the airfoil (compare Figure \ref{fig:rae2822Airfoil}). Subfigures (b), (c), and (e) present the surface pressure predictions and the CFD reference given the control input which is depicted in Subfigure (a). The shock position can be seen on the upper surface as a distinct colour change (transitioning from dark blue to light blue). As visible in Subfigure (c), the shock predicted by the GNS smears out over time. This is verified by the strong error at the shock position in Subfigure (d). In contrast, the GNODE does not smooth the shock over time, as visible in Subfigures (e) and (f). The strongest deviation of the GNODE prediction from the CFD reference can also be found at the shock. Subfigures (g)--(i) present the $c_p$-distribution at specific times $t$, depicted as dashed lines in Subfigures (a)--(f), and support the above observations. Additionally, these plots show that the GNODE framework produces significantly smoother surface predictions compared to the GNS approach, which exhibits spatial oscillations. Because the GNS operates in discrete time and is supervised only on subsequent time steps during training, small localized errors accumulate autoregressively, resulting in unphysical oscillations in the surface pressure profile. Conversely, the GNODE maintains spatially smooth pressure distributions. Because the GNODE's training loss is evaluated over entire trajectories and backpropagated directly through the numerical ODE solver, the model is inherently conditioned against generating local oscillations. Furthermore, the GNN parametrising the vector field inside the GNODE is evaluated multiple times per step according to the RK4 integration scheme. These successive derivative updates may act as a regularising mechanism that suppresses high-frequency spatial instabilities. In addition to this sample, the Appendix presents further results in \ref{subsec:additionalResults}: first, the results for the test case $\alpha_0=3.5^\circ$ and $\hat{\alpha}=2.0^\circ$ are consistent with previous observations. For this sample, GNODE's accuracy deteriorates, especially at the shock position. Furthermore, the Appendix contains plots of the skin friction coefficient $c_{f,x}$. The results are in line with the observations made for the surface pressure.

\begin{figure}
    \centering
    \includegraphics[width=0.35\textwidth]{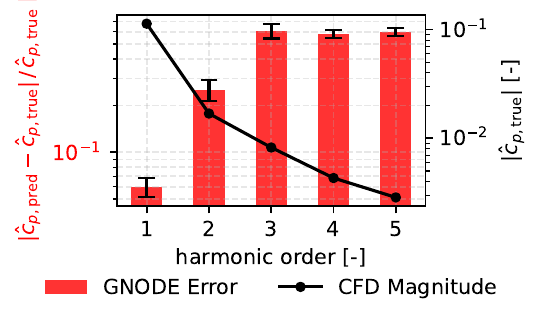}
    \caption{Red: relative error of the harmonics from a Fast Fourier Transform (FFT) of the GNODE predictions compared to the CFD reference test set. Black: magnitude of the FFT harmonics for the CFD reference test set.}
    \label{fig:globalfftrelativeerror}
\end{figure}

\begin{figure}
    \centering
    \includegraphics[width=0.55\textwidth]{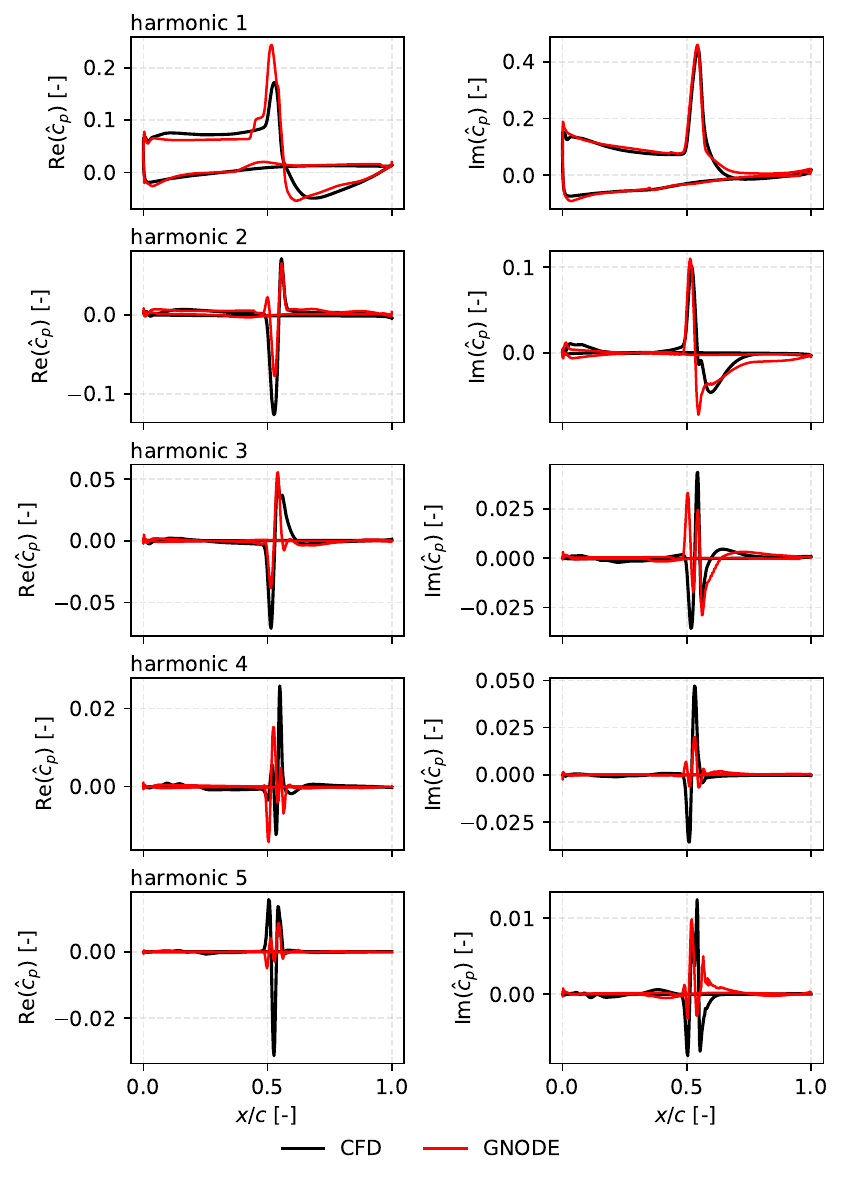}
    \caption{Comparison of the spatial distribution of the real (left column) and imaginary (right column) components of the Fourier-transformed surface pressure $\hat{c}_p$ across the first five harmonic orders for the sample $\alpha_0 = 2.79^\circ$ and $\hat{\alpha} = 1.0^\circ$.}
    \label{fig:cp_fft_Alpha2.79_Amp1.0}
\end{figure}

The time domain results provide initial insights into model performance. To gain a deeper understanding of the underlying flow physics, we conduct a frequency domain analysis on the non-transient portions of the trajectories. Since the GNS predictions exhibit significant temporal drift, non-stationary artifacts persist throughout the entire trajectory. Consequently, GNS is disregarded for this analysis. An FFT is performed on the last pitching period at the end of each CFD reference solution and its corresponding GNODE prediction. 

Figure \ref{fig:globalfftrelativeerror} shows that the amplitude of higher harmonics decays rapidly. The relative error for each harmonic predicted by GNODE increases with its order, suggesting that the model acts as a low-pass filter where high-frequency components are poorly reconstructed. To further investigate this behaviour, Figure \ref{fig:cp_fft_Alpha2.79_Amp1.0} illustrates the real and imaginary parts of the first five harmonics of the Fourier-transformed surface pressure at $\alpha_0=2.79^\circ$ and $\hat{\alpha}=1.0^\circ$. The Appendix \ref{subsec:additionalResults} contains a similar comparison for the most challenging case ($\alpha_0=3.5^\circ$, $\hat{\alpha}=2.0^\circ$). These spatial frequency plots support the previous findings: while accuracy diminishes for higher harmonics, the GNODE model correctly identifies their spatial location, i.e., the peaks still align with the shock position. However, the magnitude of these high-order peaks is not captured accurately. This frequency domain behaviour offers an explanation for the localised deviations observed at the shock position in Figure \ref{fig:surfacePlotAlpha2.79Amp1.0}, where high-frequency oscillations are present at the shock location. Due to the decreasing magnitude of the higher harmonics, compare Figure \ref{fig:globalfftrelativeerror}, the impact of these deviations on the overall results diminishes quickly.

\subsection{Augmentation}\label{subsec:augmentedDimensions}
\begin{figure}
    \centering
    \includegraphics[width=0.25\textwidth]{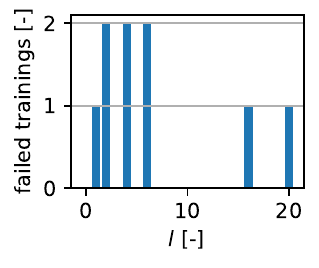}
    \caption{Number of failed training runs as a function of the number of augmented latent dimensions $l$. Each configuration was trained five times; failed attempts were retrained.}
    \label{fig:failedTrainings}
\end{figure}

The results presented above demonstrate that the GNODE approach, augmented with latent dimensions, yields a highly expressive model with good accuracy. To isolate the influence of this augmentation on performance, we conducted a study varying the number of augmented dimensions $l \in \{0, 1, 2, 4, \dots, 20\}$, while maintaining all other hyperparameters constant as defined in Subsection \ref{subsec:modelSelection}. Each model was trained five times to account for statistical variations. Figure \ref{fig:failedTrainings} shows that some training runs failed to converge; these failures occurred only when augmented dimensions were used ($l > 0$). A plausible explanation is that the latent augmentation components are not sufficiently weighted in the regularisation term. Consequently, if the loss function does not adequately constrain the augmented dynamics, the model weights may drift toward suboptimal values during training.

\begin{figure}
    \centering
    \begin{subfigure}{0.35\textwidth}
        \includegraphics[width=\textwidth]{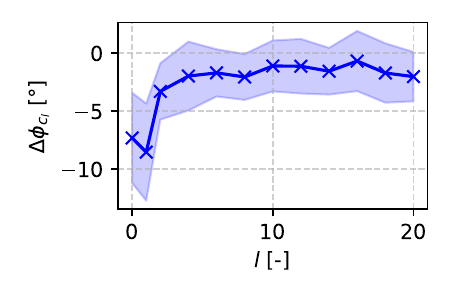}
        \caption{Lift coefficient.}
        \label{subfig:averaged_overall_coeff0}
    \end{subfigure}
    \begin{subfigure}{0.35\textwidth}
        \includegraphics[width=\textwidth]{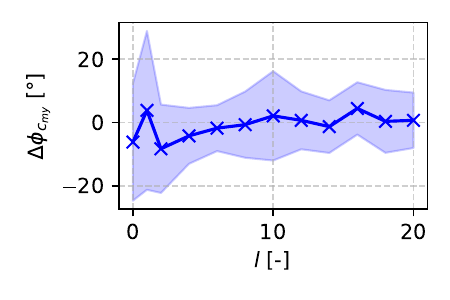}
        \caption{Pitching moment coefficient.}
        \label{subfig:averaged_overall_coeff2}
    \end{subfigure}
    \caption{Phase differences $\Delta\phi$ of the first harmonic of the last excitation period, averaged over all test samples as a function of augmented dimensions $l$. The line presents the mean values, and the shaded area represents the standard deviation.}
    \label{fig:augmentedDimsPhaseShiftAveraged}
\end{figure}

Figure \ref{fig:augmentedDimsPhaseShiftAveraged} illustrates the phase difference $\Delta\phi$ between the reference and predicted results as a function of $l$. The phase difference is computed by comparing the first harmonic of the Fourier transform applied to the last oscillation period. Increasing the number of augmented dimensions leads to reduced phase shifts, with improvements becoming more pronounced at larger values of $l$. Additionally, the variability in results (shaded area) for $l > 0$ is notably smaller than that observed without augmentation ($l=0$), suggesting that the augmented dimensions enhance prediction consistency. This behaviour can be explained by the theoretical constraints of Neural ODEs. Unaugmented Neural ODEs are restricted by the uniqueness theorem, which prohibits trajectory crossing \cite{dupont2019augmented}. However, unsteady aerodynamics of a pitching airfoil often exhibit such crossing trajectories, i.e. aerodynamic hysteresis. Lifting the problem into a higher-dimensional space by adding augmented dimensions ($l > 0$) provides additional degrees of freedom, resolving this limitation and resulting in a reduced phase lag $\Delta\phi$.

\section{Conclusions}\label{sec:conclusions}
We introduce the Graph Neural Ordinary Differential Equation (GNODE) framework for predicting unsteady aerodynamic fields with external control inputs. GNODE utilises message passing in a discretised spatial domain to parametrise flow dynamics and employs Neural ODEs for temporal propagation, making the method particularly suited for spatio-temporal modelling of systems undergoing forced motions. To overcome the expressivity limitations of standard Neural ODEs, we further enhance the architecture by augmenting the dynamics vector with latent dimensions, that help capturing history effects. To validate this approach, we developed a high-fidelity dataset of URANS simulations for a periodically pitching RAE 2822 airfoil under transonic conditions using a full-factorial design of experiments. The dataset comprises 40 trajectories exhibiting both linear and non-linear dynamic behaviour with moving shocks; models were trained on 32 trajectories and evaluated on the remaining eight at a test frequency of $k=0.3$ against an autoregressive discrete-time GNS baseline.

Our results demonstrate that GNODE consistently outperforms the GNS baseline across multiple metrics, particularly in predicting surface quantities and integrated global coefficients. Unlike the baseline, which suffers from autoregressive error accumulation and temporal drift, GNODE produces stable roll-outs while capturing aerodynamic shocks with significantly less smearing, despite both models facing challenges with high-frequency oscillations. We further isolate the effect of latent dimension augmentation, showing that it systematically increases prediction accuracy and reduces phase shift relative to the reference solution. The framework also showcases robustness to random initialisations. However, several limitations must be acknowledged. First, accuracy degrades toward the edges of the design space due to the basic full-factorial DoE, particularly in the transonic regime where moving shocks induce severe dynamic non-linearities. Second, GNODE demands significantly higher computational effort and longer training times compared to the discrete-time GNS. Finally, while inference is statistically robust, training can be unstable, potentially because the regularisation of the augmented latent dimensions is insufficient to fully constrain their evolution.

To address these limitations, future research should first focus on data and implementation improvements. Transitioning from sparse full-factorial designs to quasi-random sampling techniques, such as Latin Hypercube or Sobol sampling \cite{sobol1999use, tang1993orthogonal}, could increase information density, while employing Schroeder multi-sine signals \cite{schroeder1970synthesis} instead of multiple sinusoids may further improve data efficiency. The current single-batch training restriction, which requires uniform time steps, could be relaxed through architectural modifications that accommodate variable sampling rates or an a-priori data processing. Building on these improvements, extensions to more complex physical scenarios are possible. This includes applying the model to combined pitch-and-heave motions or transient effects like gusts, and scaling to three-dimensional aerodynamics by incorporating spatially varying control signals to handle geometric deformations. Extending the framework to interpolate discrete control inputs, rather than relying on analytic functions, would also broaden practical applicability at the expense of potential approximation errors. From a theoretical standpoint, the interpretability of latent dimensions warrants investigation to determine whether they correlate with unmodelled physical quantities or if the model differentiability can be leveraged for sensitivity analysis and aerodynamic optimisation. Finally, the numerical implementation should advance toward adaptive-step or implicit integration schemes \cite{fronk2024training}, which are essential for mathematically stiff transonic regimes. Ultimately, given its domain-agnostic nature, GNODE could serve as a generalised framework for spatio-temporal systems across various other engineering disciplines.

\newpage
\printcredits

\section*{Declaration of Competing Interest}
The authors declare that they have no known competing financial interests or personal relationships that could have appeared to influence the work reported in this paper.

\section*{Acknowledgements}
The authors gratefully acknowledge the scientific support and HPC resources provided by the German Aerospace Center (DLR). The HPC system CARA is partially funded by “Saxon State Ministry for Economic Affairs, Labour and Transport” and “Federal Ministry for Economic Affairs and Energy”.

\section*{Declaration of Generative AI and AI-assisted Technologies in the Manuscript Preparation Process}
During the preparation of this work the authors used the large language model Ollama (DLR internal) in order to assist with text restructuring and phrasing. After using this tool/service, the authors reviewed and edited the content as needed and take full responsibility for the content of the published article.

\section*{Data Availability Statement}
Data will be made available from the corresponding author upon reasonable request.

\appendix
\section{Appendix}\label{sec:appendix}
\subsection{CFD Simulation Details}\label{subsec:CFDSimulationDetails}
Table \ref{tab:timeStepSizes} shows the step sizes that are selected based on time step convergence study. The number of periods, i.e. of the periodic pitch, is selected such that the first harmonic of the global coefficient are converging.
\begin{table}
    \centering
    \begin{tabular}{c|c|c|c|c}
         $k$ [-] & $\omega$ [s$^{-1}$] & $\Delta t$ [$\cdot10^{-3}$\,s] & steps per period [-] & periods [-] \\ \hline
         $0.1$ & $24.2$ & $1.01$ & $256$ & $2$ \\
         $0.2$ & $48.4$ & $1.01$ & $128$ & $3$ \\
         $0.3$ & $72.6$ & $1.35$ & $64$ & $5$ \\
         $0.5$ & $120.9$ & $0.81$ & $64$ & $5$ \\
         $1.0$ & $241.8$ & $0.81$ & $32$ & $10$ \\
    \end{tabular}
    \caption{Parameters for the time stepping of the unsteady CFD simulations.}
    \label{tab:timeStepSizes}
\end{table}

\subsection{Metrics}\label{subsec:metrics}
\begin{figure}
    \centering
    \begin{subfigure}{0.6\textwidth}
        \includegraphics[width=\textwidth]{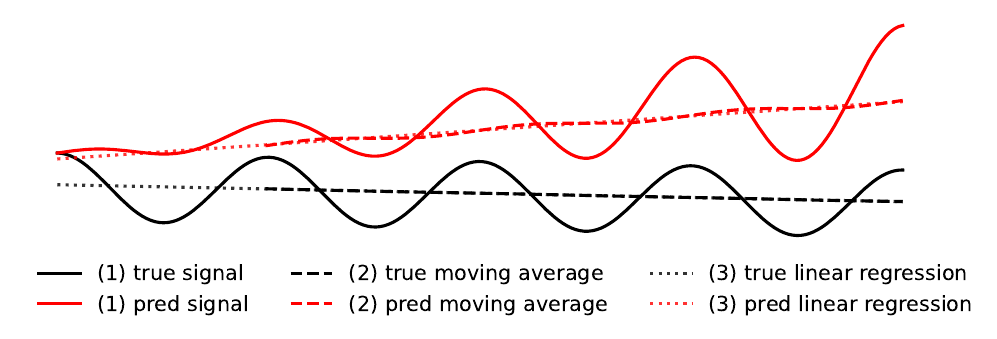}
        \caption{Drift: the solid lines (1) show exemplary true and predicted signals. In a first step a moving average given the true signals and number of steps per period is performed. This is shown as dashed lines (2). Secondly, linear regressions are fitted through the moving averages (see the dotted lines (3)). To compute the drift, the slopes of the linear regressions are compared as in Equation \ref{eq:drift}.}
        \label{subfig:drift}
    \end{subfigure}\\
    \begin{subfigure}[t]{0.48\textwidth}
        \includegraphics[width=\textwidth]{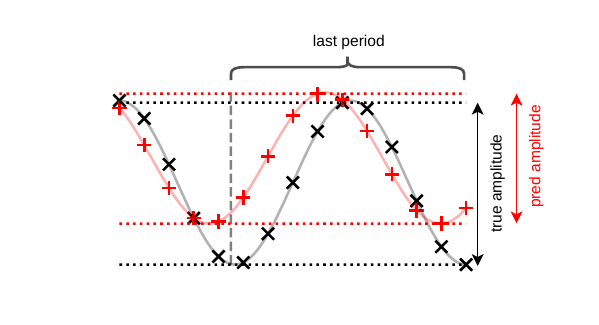}
        \caption{Amplitude deviation: the minimal and maximal values of the true and the predicted signal within the last period are used to compute the amplitudes. The corresponding metric is then computed according to Equation \ref{eq:amplitudeDeviation}.}
        \label{subfig:amplitudeDeviation}
    \end{subfigure}\hfill
    \begin{subfigure}[t]{0.48\textwidth}
        \includegraphics[width=\textwidth]{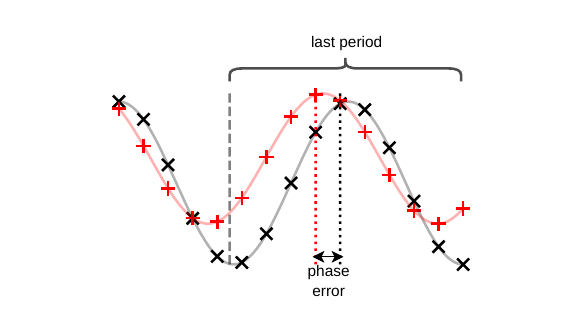}
        \caption{Phase error: the difference between the maximum values of the true and predicted signals within the last period are compared to compute the phase error according to Equation \ref{eq:phaseError}.}
        \label{subfig:phaseError}
    \end{subfigure}
    \caption{Visualisations of drift, amplitude deviation and phase error metrics.}
    \label{fig:metrics}
\end{figure}

The direct outputs of the models are local surface quantities ($c_p$, $c_{f,x}$, $c_{f,z}$). The accuracy of these surface predictions is evaluated using three standard statistical metrics:
\begin{itemize}
    \item \textbf{mean absolute error (MAE):} measures the average magnitude of the errors:
    \begin{equation}
        \text{MAE} = \frac{1}{N}\sum^N_{i=1}|y_i-\hat{y}_i|
    \end{equation}
    where $y_i$ is the true value, $\hat{y}_i$ is the predicted value, and $N$ represents the total number of evaluation points.
    \item \textbf{mean squared error (MSE):} quantifies the average of the squared differences:
    \begin{equation}
        \text{MSE} = \frac{1}{N}\sum^N_{i=1}(y_i-\hat{y}_i)^2
    \end{equation}
    In contrast to MAE, this metric penalises larger deviations stronger.
    \item \textbf{coefficient of determination ($R^2$):} measures the proportion of variance in the reference data that is captured by the model:
    \begin{equation}
        R^2 = 1 - \frac{\sum^N_{i=1}(y_i-\hat{y}_i)^2}{\sum^N_{i=1}(y_i-\bar{y}_i)^2}
    \end{equation}
    where $\bar{y} = \frac{1}{N}\sum^N_{i=1}y_i$ represents the mean value of the true dataset.
\end{itemize}
Global aerodynamic coefficients ($c_l$, $c_d$, $c_{my}$) computed by taking the surface integral of the local forces ($c_p$, $c_{f,x}$, $c_{f,z}$). This work focuses on the pitching moment coefficient. In unsteady aerodynamics, $c_{my}$ usually is the most sensitive global coefficient. In addition to MAE, MSE, and $R^2$ the global quantities are compared on three additional metrics:
\begin{itemize}
    \item \textbf{drift:} quantifies the distributional shift of autoregressive roll-outs. First, a moving average of the true and predicted signals is calculated over an oscillation period:
    \begin{equation}
        \bar{y}_i = \frac{1}{\mathit{stepsPerPeriod}}\sum^{\mathit{stepsPerPeriod}-1}_{j=0}y_{i+j}
    \end{equation}
    \begin{equation}
        \bar{\hat{y}}_i = \frac{1}{\mathit{stepsPerPeriod}}\sum^{\mathit{stepsPerPeriod}-1}_{j=0}\hat{y}_{i+j}
    \end{equation}
    for indices $i \in \{1, \dots, \mathit{signalLength} - \mathit{stepsPerPeriod} + 1\}$. A linear regression is then fit through the moving average outputs. The drift rate compares the slope of the prediction to the slope of the reference data:
    \begin{equation}\label{eq:drift}
        \mathit{drift} = |m_{\text{pred}} - m_{\text{true}}|
    \end{equation}
    where $m_{\text{pred}}$ and $m_{\text{true}}$ are the slopes of linear regression through the predicted and reference moving averages, respectively.
    \item \textbf{amplitude deviation:} measures whether a model dampens or amplifies the oscillatory aerodynamic response over time. The peak-to-peak amplitudes are calculated looking at the last pitching period:
    \begin{equation}
        A_{\text{true}} = \max_{i \in \mathit{lastPeriod}}(y_i) - \min_{i \in \mathit{lastPeriod}}(y_i)
    \end{equation}
    \begin{equation}
        A_{\text{pred}} = \max_{i \in \mathit{lastPeriod}}(\hat{y}_i) - \min_{i \in \mathit{lastPeriod}}(\hat{y}_i)
    \end{equation}
    The non-dimensional amplitude deviation is defined as:
    \begin{equation}\label{eq:amplitudeDeviation}
        \mathit{amplitude\ deviation} = \left| \frac{A_{\text{pred}}}{A_{\text{true}}} - 1 \right|
    \end{equation}
    \item \textbf{phase error:} determines whether the model correctly preserves the phase the physics. The discrete time steps corresponding to the maximum peak locations during the final oscillation period are identified:
    \begin{equation}
        i_{\text{true}} = \arg\max_{i \in \mathit{lastPeriod}}(y_i)
    \end{equation}
    \begin{equation}
        i_{\text{pred}} = \arg\max_{i \in \mathit{lastPeriod}}(\hat{y}_i)
    \end{equation}
    The angular phase error in degrees is then computed as:
    \begin{equation}\label{eq:phaseError}
        \mathit{phase\ error} = \left( \frac{|i_{\text{pred}} - i_{\text{true}}|}{\mathit{stepsPerPeriod}} \right) \cdot 360^\circ
    \end{equation}
\end{itemize}
Figure \ref{fig:metrics} depicts exemplary visualisations of the drift, amplitude deviation and phase error metrics. Note that the resolution of the signals, i.e. the sampling rate, influences the outcome of the metrics.

\subsection{Additional Results}\label{subsec:additionalResults}
\begin{figure}
\centering
\begin{subfigure}{0.32\textwidth}
    \includegraphics[width=\textwidth]{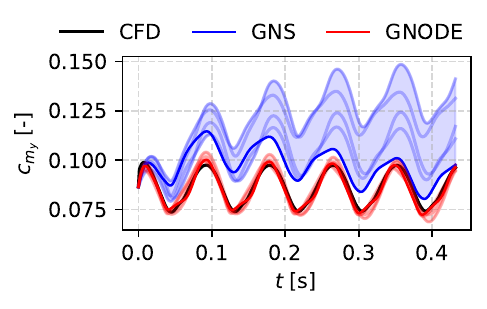}
    \caption{$\alpha_0=0.0^\circ$, $\hat\alpha=2.0^\circ$.}
    \label{subfig:coeff2Alpha0.0Amp2.0}
\end{subfigure}
\begin{subfigure}{0.32\textwidth}
    \includegraphics[width=\textwidth]{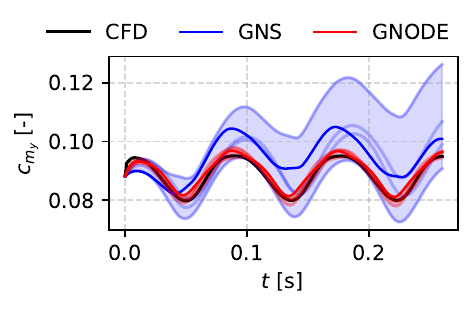}
    \caption{$\alpha_0=1.5^\circ$, $\hat\alpha=1.0^\circ$.}
    \label{subfig:coeff2Alpha1.5Amp1.0}
\end{subfigure}
\begin{subfigure}{0.32\textwidth}
    \includegraphics[width=\textwidth]{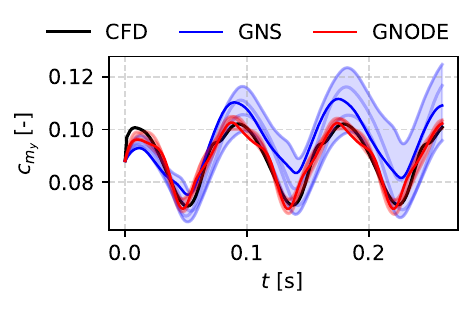}
    \caption{$\alpha_0=1.5^\circ$, $\hat\alpha=2.0^\circ$.}
    \label{subfig:coeff2Alpha1.5Amp2.0}
\end{subfigure}
\begin{subfigure}{0.32\textwidth}
    \includegraphics[width=\textwidth]{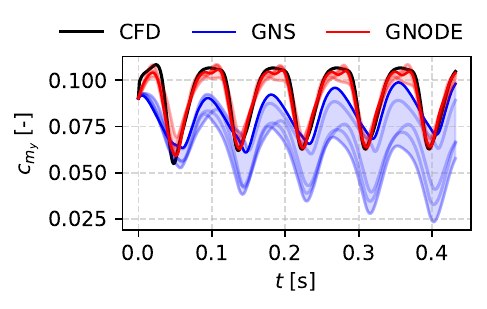}
    \caption{$\alpha_0=2.79^\circ$, $\hat\alpha=2.0^\circ$.}
    \label{subfig:coeff2Alpha2.79Amp2.0}
\end{subfigure}
\begin{subfigure}{0.32\textwidth}
    \includegraphics[width=\textwidth]{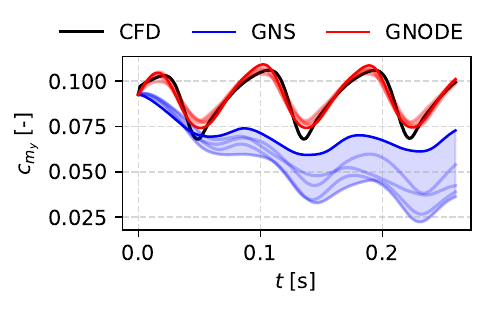}
    \caption{$\alpha_0=3.5^\circ$, $\hat\alpha=1.0^\circ$.}
    \label{subfig:coeff2Alpha3.5Amp1.0}
\end{subfigure}
\caption{Pitching moment coefficient $c_{my}$ over time $t$. $c_{my}$ is calculated using $c_p$, $c_{f,x}$ and $c_{f,z}$. For GNS and GNODE the dark-coloured lines correspond to the best of five models according to the metrics in table \ref{tab:metricsFullDatasetTest}. The light-coloured lines show the remaining four models. The shaded regions indicate the spread of predictions across the model instances.}
\label{fig:globalCoefficientsAppendix}
\end{figure}

\begin{figure}
    \centering
    \includegraphics[width=0.75\textwidth]{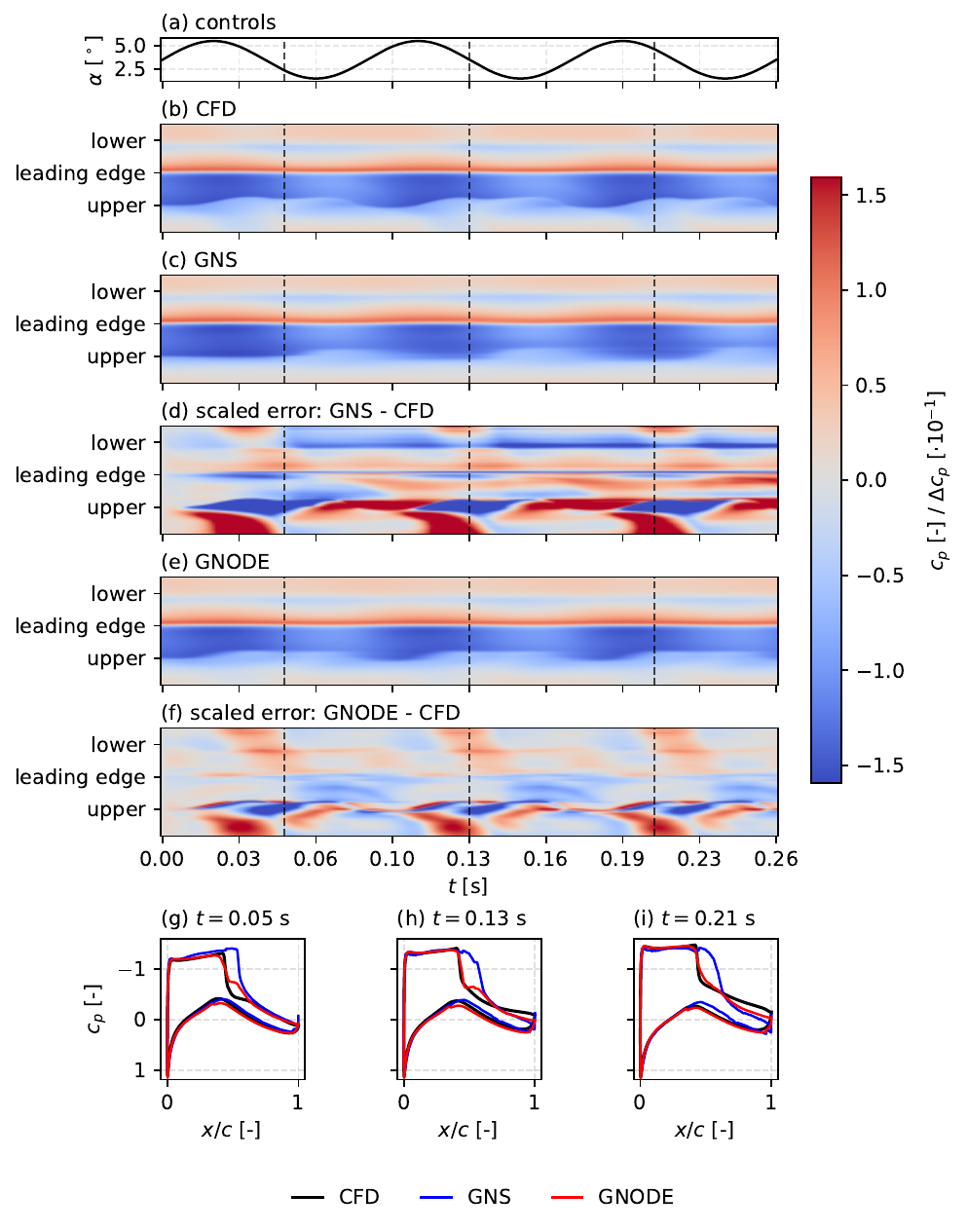}
    \caption{Spatio-temporal evolution of the surface quantities for $\alpha_0=3.5^\circ$, $\hat{\alpha}=2.0^\circ$. Panel (a) shows $\alpha$, which is part of the forcing control signal. Subfigures (b)--(f) show the evolution of the surface pressure over time: The y-axis corresponds to the \textit{unrolled} surface of the airfoil (compare Figure \ref{fig:rae2822Airfoil}). Panel (b) shows the CFD reference, while (c) and (e) show the predictions of the best GNS and GNODE models according to the metrics in Table \ref{tab:metricsFullDatasetTest}. Panels (d) and (f) show the error relative to the CFD reference solution. Subfigures (g), (h), and (i) present $c_p$-distributions along the length of the airfoil ($x/c$) at three times $t$. To highlight the errors, the values (and colours) are amplified by a factor of 10. The vertical dashed lines in (a)--(f) present the times $t$ corresponding to Subfigures (g)--(i).}
    \label{fig:surfacePlotAlpha3.5Amp2.0}
\end{figure}

\begin{figure}
    \centering
    \includegraphics[width=0.75\textwidth]{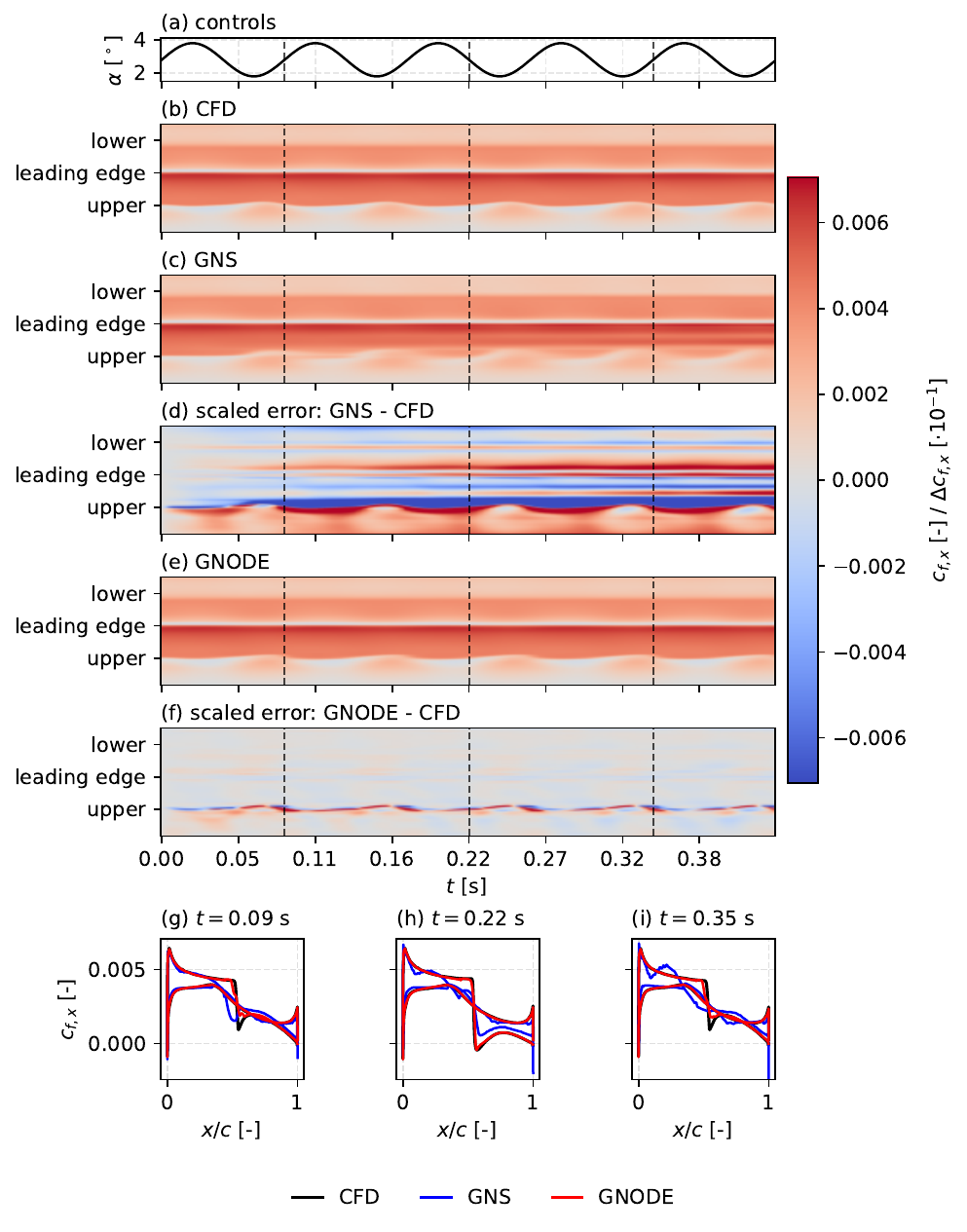}
    \caption{Spatio-temporal evolution of the surface quantities for $\alpha_0=2.79^\circ$, $\hat{\alpha}=1.0^\circ$. Panel (a) shows $\alpha$, which is part of the forcing control signal. Subfigures (b)--(f) show the evolution of the skin friction coefficient over time: The y-axis corresponds to the \textit{unrolled} surface of the airfoil (compare Figure \ref{fig:rae2822Airfoil}). Panel (b) shows the CFD reference, while (c) and (e) show the predictions of the best GNS and GNODE models according to the metrics in Table \ref{tab:metricsFullDatasetTest}. Panels (d) and (f) show the error relative to the CFD reference solution. Subfigures (g), (h), and (i) present $c_{f,x}$-distributions along the length of the airfoil ($x/c$) at three times $t$. To highlight the errors, the values (and colours) are amplified by a factor of 10. The vertical dashed lines in (a)--(f) present the times $t$ corresponding to Subfigures (g)--(i).}
    \label{fig:surfacePlocfAlpha2.79Amp1.0}
\end{figure}

\begin{figure}
    \centering
    \includegraphics[width=0.75\textwidth]{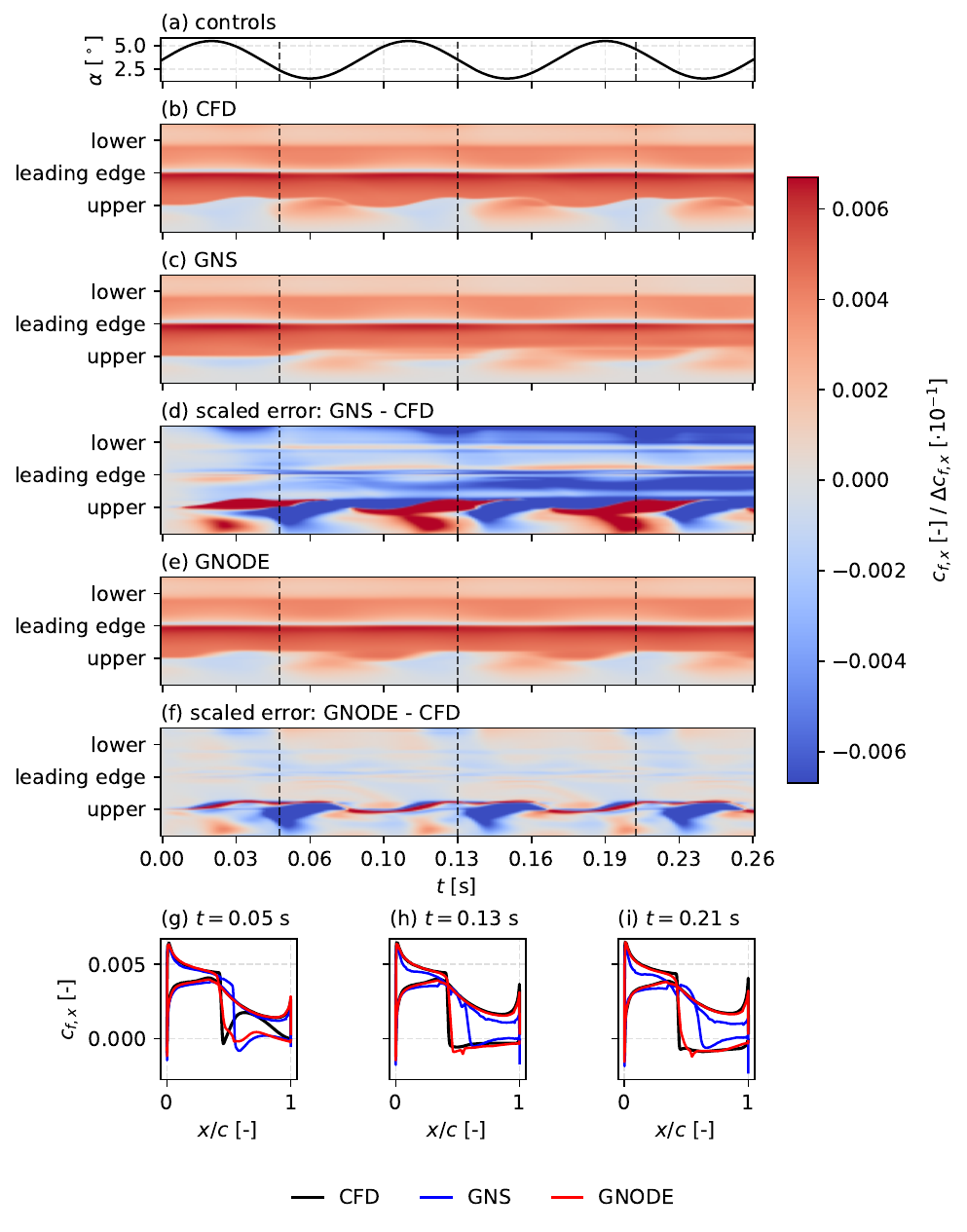}
    \caption{Spatio-temporal evolution of the surface quantities for $\alpha_0=3.5^\circ$, $\hat{\alpha}=2.0^\circ$. Panel (a) shows $\alpha$, which is part of the forcing control signal. Subfigures (b)--(f) show the evolution of the skin friction coefficient over time: The y-axis corresponds to the \textit{unrolled} surface of the airfoil (compare Figure \ref{fig:rae2822Airfoil}). Panel (b) shows the CFD reference, while (c) and (e) show the predictions of the best GNS and GNODE models according to the metrics in Table \ref{tab:metricsFullDatasetTest}. Panels (d) and (f) show the error relative to the CFD reference solution. Subfigures (g), (h), and (i) present $c_{f,x}$-distributions along the length of the airfoil ($x/c$) at three times $t$. To highlight the errors, the values (and colours) are amplified by a factor of 10. The vertical dashed lines in (a)--(f) present the times $t$ corresponding to Subfigures (g)--(i).}
    \label{fig:surfacePlotcfAlpha3.5Amp2.0}
\end{figure}

\begin{figure}
    \centering
    \includegraphics[width=0.6\textwidth]{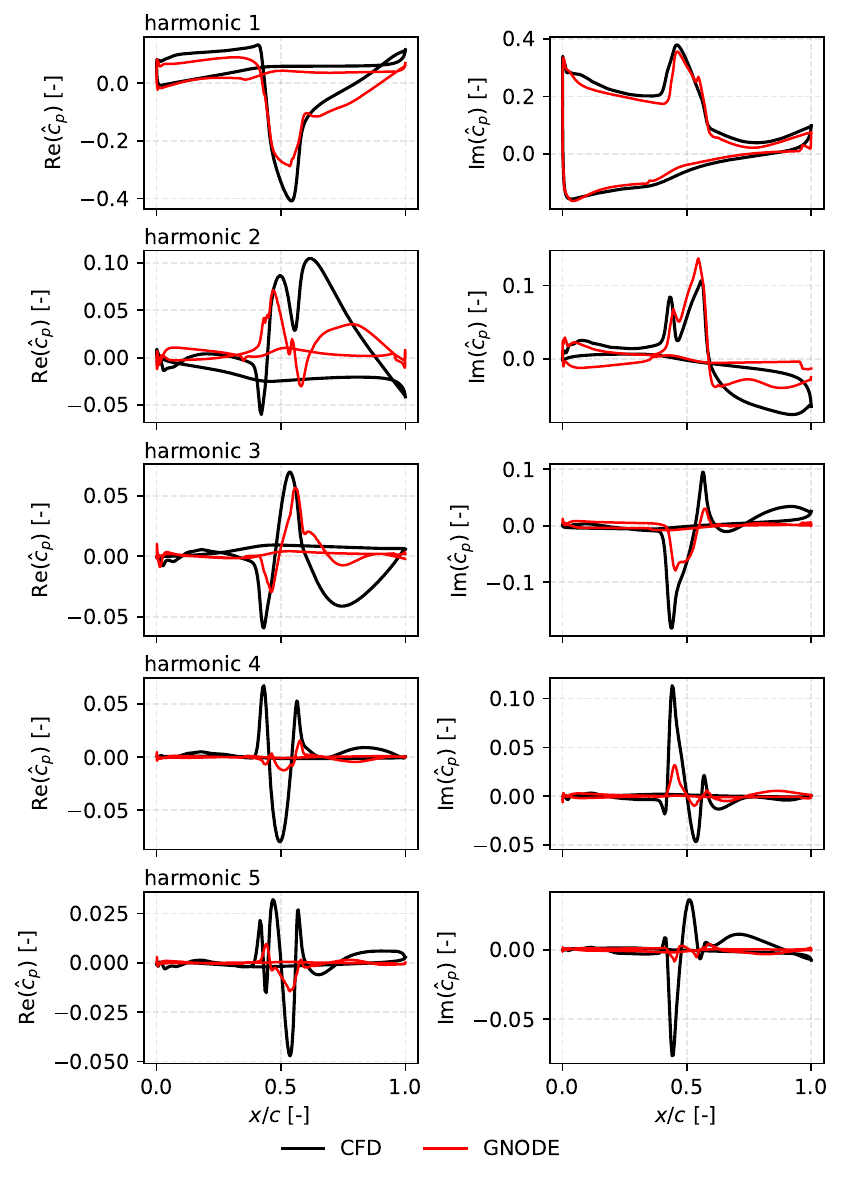}
    \caption{Comparison of the spatial distribution of the real (left column) and imaginary (right column) components of the Fourier-transformed surface pressure coefficient $\hat{c}_p$ across the first five harmonic orders for the sample $\alpha_0 = 3.5^\circ$ and $\hat{\alpha} = 2.0^\circ$).}
    \label{fig:cp_fft_Alpha3.5_Amp2.0}
\end{figure}

Figure \ref{fig:globalCoefficientsAppendix} presents complementary results to Figure \ref{fig:globalCoefficients}.
Figures \ref{fig:surfacePlotAlpha3.5Amp2.0} to \ref{fig:surfacePlotcfAlpha3.5Amp2.0} present complementary results to Figure \ref{fig:surfacePlotAlpha2.79Amp1.0}. Figure \ref{fig:cp_fft_Alpha3.5_Amp2.0} presents complementary results to Figure \ref{fig:cp_fft_Alpha2.79_Amp1.0}. The findings from Section \ref{sec:results} are supported by the additional results: GNODE produces consistently more accurate results with a smaller variance compared to GNS.

\FloatBarrier

\bibliographystyle{model1-num-names}

\bibliography{cas-refs}



\end{document}